\documentclass[twoside]{IEEEtran}
\usepackage{graphicx,amssymb,amsmath,multirow,float,cite,subfigure,CJK,booktabs} %,subcaption，caption

\ifCLASSINFOpdf
\else
\fi
\begin{document}
\title{Stroke-Based Scene Text Erasing \\ Using Synthetic Data for Training}
% a structure at a ~ so this keeps an author', s name from being broken across
% a separate \thanks must be used for each paragraph as LaTeX2e', s \thanks

\author{Zhengmi~Tang,
        Tomo~Miyazaki,~\IEEEmembership{Member,~IEEE,}
        Yoshihiro~Sugaya,~\IEEEmembership{Member,~IEEE,}
        and~Shinichiro~Omachi,~\IEEEmembership{Senior Member,~IEEE}% <-this % stops a space

% \author{Michael~Shell,~\IEEEmembership{Member,~IEEE,}
%         John~Doe,~\IEEEmembership{Fellow,~OSA,}
%         and~Jane~Doe,~\IEEEmembership{Life~Fellow,~IEEE}% <-this % stops a space

\thanks{Manuscript received March 10, 2021. This work was partially supported by JSPS KAKENHI Grant Numbers 18K19772, 19K12033, 20H04201.}

\thanks{The authors are with the Graduate School of Engineering, Tohoku University, Sendai, 980-8579, Japan. (e-mail: tzm@dc.tohoku.ac.jp, tomo@tohoku.ac.jp, sugaya@iic.ecei.tohoku.ac.jp, machi@ecei.tohoku.ac.jp).}
}% <-this % stops a space
% \thanks{J. Doe and J. Doe are with Anonymous University.}% <-this % stops a space
% \thanks{Manuscript received April 19, 2021; revised August 26, 2015.}


\markboth{IEEE Transactions on Image Processing}% ,~Vol.~??, No.~??, ??~2021
{Tang \MakeLowercase{\textit{\textit{et al.}}}: Stroke-Based Scene Text Erasing Using Synthetic Data for Training}
% {Shell \MakeLowercase{\textit{\textit{et al.}}}: Bare Demo of IEEEtran.cls for IEEE Journals}
% *** Note that you probably will NOT want to include the author', s ***

% If you want to put a publisher', s ID mark on the page you can do it like


\maketitle

\begin{abstract}
Scene text erasing, which replaces text regions with reasonable content in natural images, has drawn significant attention in the computer vision community in recent years. There are two potential subtasks in scene text erasing: text detection and image inpainting. Both subtasks require considerable data to achieve better performance; however, the lack of a large-scale real-world scene-text removal dataset does not allow existing methods to realize their potential. To compensate for the lack of pairwise real-world data, we made considerable use of synthetic text after additional enhancement and subsequently trained our model only on the dataset generated by the improved synthetic text engine. Our proposed network contains a stroke mask prediction module and background inpainting module that can extract the text stroke as a relatively small hole from the cropped text image to maintain more background content for better inpainting results. This model can partially erase text instances in a scene image with a bounding box or work with an existing scene-text detector for automatic scene text erasing. The experimental results from the qualitative and quantitative evaluation on the SCUT-Syn, ICDAR2013, and SCUT-EnsText datasets demonstrate that our method significantly outperforms existing state-of-the-art methods even when they are trained on real-world data.
\end{abstract}


% \fi
\IEEEpeerreviewmaketitle

\section{Introduction}

\IEEEPARstart{T}{exts} created by humankind comprise rich, precise high-level semantics. Text, in daily life, provides a considerable amount of valuable information. However, with the increasing number of portable devices, such as digital cameras, tablets, smartphones, and SNS, a huge volume of scene images, including text, are shared on the Internet every second. These texts can contain private information, such as names, addresses, and vehicle number plates. With the increasing development of scene text detection and recognition technology, there is a high risk that the information collected automatically is used for illegal purposes. Therefore, scene text erasing, which is replacing text regions in scene images with proper content, has drawn considerable attention in the computer vision community in recent years. 

After the prior work of Scene Text Eraser \cite{Nakamura2017}, scene text erasing research has developed in two directions: one-step and two-step methods. One-step methods \cite{TursunMTRNetpp2020,zhang_Ensnet_2019,LiuEraseNet2020} do not require the input of any text location information because they combine text detection and inpainting functions into one network, making them lightweight and fast. The drawback is that the text localization mechanism of these networks is weak, and the text-erasing process is not controllable. To allow the network to learn the complicated distribution of scene texts, a considerable number of manually text-erased real-world images are required as training data because the text distribution of the Synth-text \cite{gupta_synthetic_2016} dataset, which is generated according to human rules, is significantly different from real-world cases. To generate scene-text-erased images \cite{LiuEraseNet2020,Biancascaded2020}, photo editing software such as Photoshop is used to fill the text region of natural images with visually plausible content. However, this type of annotation is expensive and time-consuming because the annotators need to operate carefully to guarantee the quality of the erasing, especially when the text to be erased is on a complicated background. 

\begin{figure}[!t]
\centering
\begin{minipage}[t]{0.33\linewidth}
    \centering
    \includegraphics[height=0.85\textwidth, width=1\textwidth, trim=0 38 0 38, clip ]{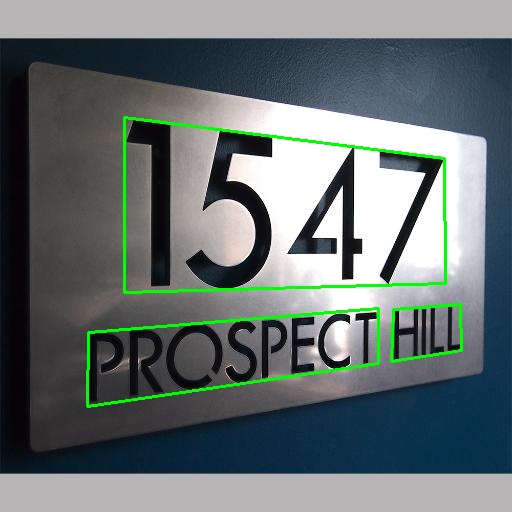}\vspace{1.5 pt}
    \includegraphics[height=0.85\textwidth, width=1\textwidth]{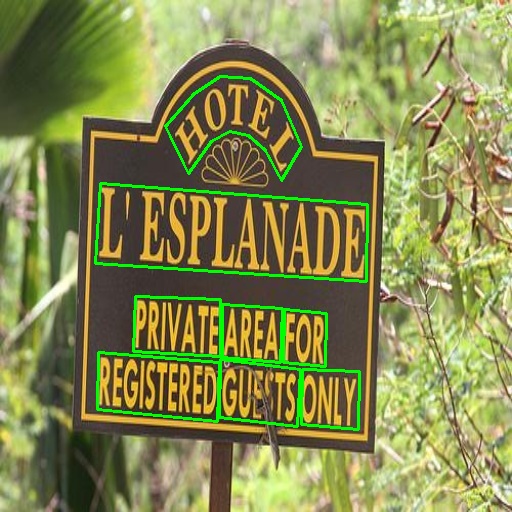}
\end{minipage}
\hspace*{-5pt}
\begin{minipage}[t]{0.33\linewidth}
    \centering
    \includegraphics[height=0.85\textwidth, width=1\textwidth, trim=0 38 0 38, clip ]{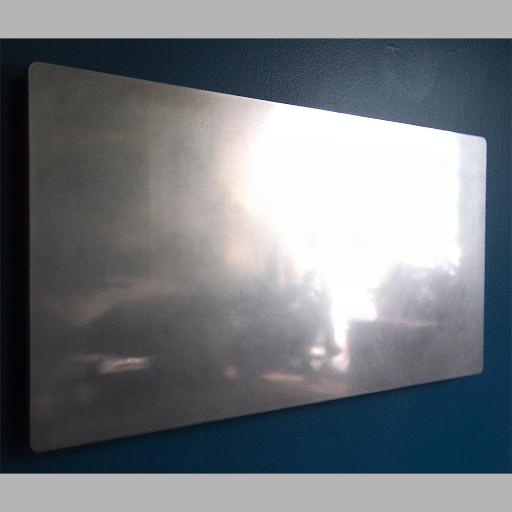}\vspace{1.5 pt}
    \includegraphics[height=0.85\textwidth, width=1\textwidth]{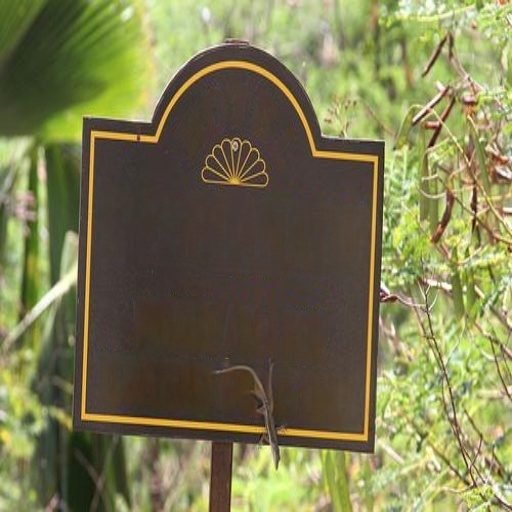}
\end{minipage}
\hspace*{-5pt}
\begin{minipage}[t]{0.33\linewidth}
    \centering
    \setlength{\fboxrule}{0.4pt}
    \setlength{\fboxsep}{0pt}
    \fbox{\includegraphics[height=0.85\textwidth, width=1\textwidth, trim=0 38 0 38, clip ]{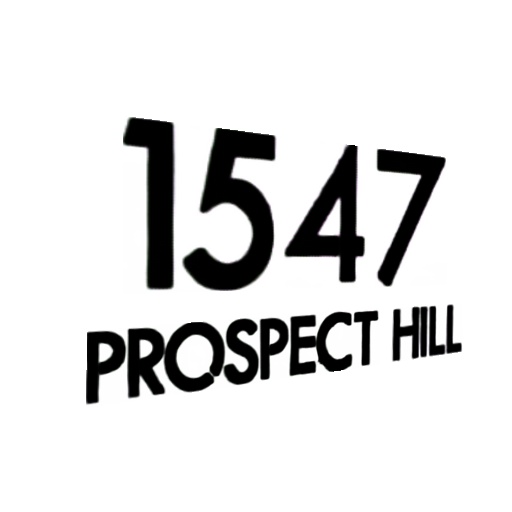}}\vspace{1.2 pt}
    \fbox{\includegraphics[height=0.85\textwidth, width=1\textwidth]{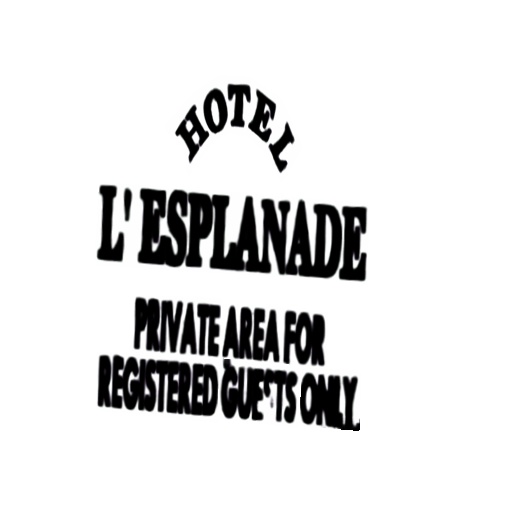}}
\end{minipage}
\caption{Scene Text Erasing: original images and text bounding boxes (left); text-erased images by our method (middle); predicted text stroke mask (right).} \label{tang1}
\end{figure}
% %
% \begin{figure}[!t]
% \includegraphics[height=0.49\textwidth ,width=0.49\textwidth]{tang1.png}
% \caption{Scene Text Erasing: original images and text bounding boxes (left); text-erased images by our method (middle); predicted text stroke mask (right).} \label{tang1}
% \end{figure}
%
The two-step approaches decompose the text-erasing task into two sub-problems: text detection and background inpainting. MTRNet \cite{TursunMTRNet2019} inpaints the text region by manually providing a mask of the text regions. Zdenek \textit{et al.} \cite{Zdenekweaksupervision2020} used a pretrained scene text detection model and a pretrained inpainting model to erase the text. This weak supervision method does not require paired training data. However, the inpainting model is trained on the Street View \cite{DoerschParisstreetview2012} or ImageNet \cite{imageNet2015} datasets, and thus, the pretrained models face domain shift problems, which cannot make a “perfect fit" in the context of scene text.

In this study, we propose a word-level two-stage scene-text-erasing network that works on cropped text images. First, it predicts the region of text stroke as a hole, then inpaints the hole according to the background. To compensate for the lack of pairwise real-world data, we made full use of synthetic texts. The appearance of the synthetic texts \cite{gupta_synthetic_2016} was enhanced, and we trained our model only on the dataset generated by the improved synthetic text engine in an end-to-end fashion. The model can partially erase text instances in a scene image if text bounding boxes are provided. The model can also work with existing scene-text detectors for automatic scene text erasing. Examples of text-erasing results obtained using the proposed method are shown in Fig.~\ref{tang1}.

The main contributions of our study can be summarized as follows:
\begin{itemize}
\item[\textbullet ] We propose a practical text erasing method and a stroke-based text-erasing network using cropped text images instead of the entire image, which makes prediction of the pixel-level mask of text strokes more accurate and stable. Benefiting from our erasing pipeline and network structure, our method can erase text instances while retaining and restoring more background details.

\item[\textbullet ] We enhance the Text Synthesis Engine \cite{gupta_synthetic_2016} to make the appearance of synthetic text instances share more similarity with real-world data.

\item[\textbullet ] The quantitative and qualitative evaluation results on SCUT-Syn \cite{zhang_Ensnet_2019}, ICDAR 2013 \cite{KaratzasICDAR2013} and SCUT-EnsText \cite{LiuEraseNet2020} datasets demonstrate that our method outperforms previous state-of-the-art methods while it is only trained on the dataset generated by the improved synthetic text engine \cite{gupta_synthetic_2016}.

\end{itemize}

The remainder of this paper is organized as follows. Section \ref{Related work} reviews related works on scene-text detection, image inpainting, and text erasing. Section \ref{Methodology} introduces the details of our method, including the pipeline and the proposed networks. In Section \ref{Experiment}, we evaluate and compare our proposed method with related inpainting and scene-text erasing studies based on the results of experiments. Finally, we provide concluding statements in Section \ref{Conclusion}.

%figure2
\begin{figure*}[!t]
\centering
\includegraphics[width=\textwidth]{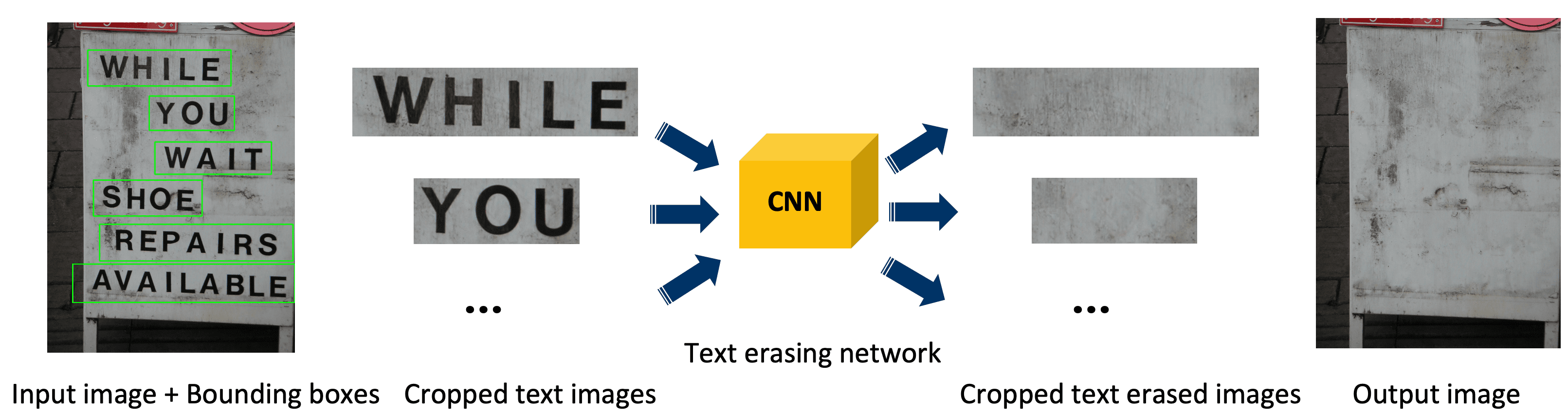}
\caption{Pipeline of our proposed method. Our text-erasing network uses cropped text images as inputs and outputs the corresponding text-erased images. Then, by placing these text-erased images back into the original locations of the input image, we can obtain the final text-erased image.} \label{tang2}
\end{figure*}
%

%figure3
\begin{figure*}[!t]
\centering
\includegraphics[width=0.96\textwidth]{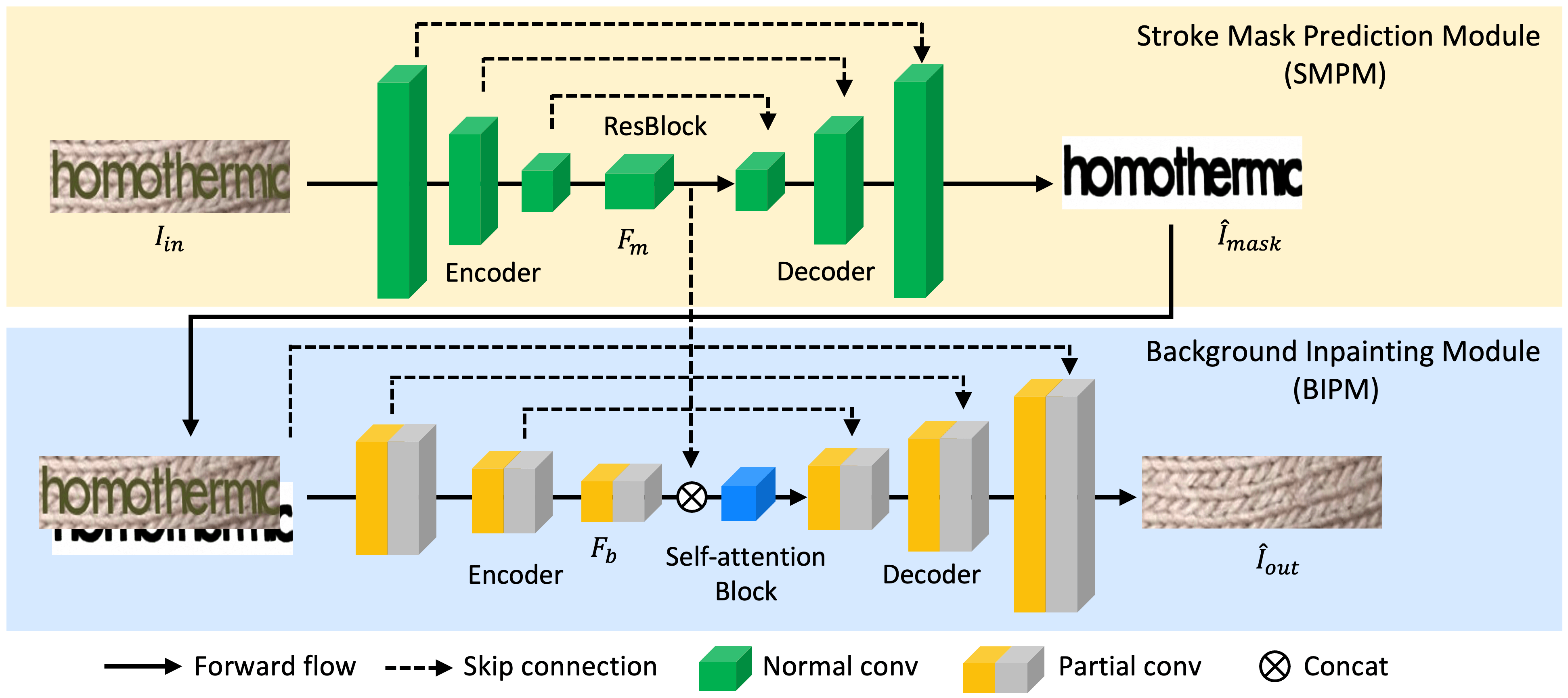}
\caption{Structure of our proposed network. It is composed of a stroke mask prediction module (SMPM) (top) and a background inpainting module (BIPM) (bottom). The SMPM first predicts the stroke mask of the text, and the BIPM inpaints the stroke pixels with the appropriate content to erase the text.} \label{tang3}
\end{figure*}

\section{Related work} %2
\label{Related work}
\subsection{Scene Text Detection} % 2.1
The emergence of deep learning has facilitated the development of scene text detection research and shows promising performance compared to traditional manually designed algorithms \cite{chen_textnaturalscenes_2004,Neumann_textlocandrec_2011,jamil_edge-based_2011,mosleh_automatic_2013,huang_textloc_2013}. Recent learning-based scene text detection methods can be roughly categorized into regression-based and segmentation-based methods.
Regression-based methods aim to directly predict the bounding boxes of text instances. TextBoxes \cite{Liaotextbox2017} adjusts the aspect ratios of anchors in SSD \cite{liu_ssd_2016} to detect text with different shapes. CTPN \cite{tian_CTPN_2016} combines the framework of Faster R-CNN \cite{RenfasterRCNN2015} with a recurrence mechanism to predict the contextual and dense components of text. RRPN \cite{ma_RRPN_2018} proposes a rotation region proposal to bind multi-oriented scene text with rotated rectangles. EAST \cite{zhou_east_2017} directly regresses rotated rectangles or quadrangles of text through a simplified pipeline without using any anchors.
LOMO \cite{zhang_look_2019} detects long text and arbitrarily shaped text in scene images by refining the preliminary proposals iteratively and considering text geometry properties including text region, text center line, and border offsets.

Segmentation-based methods usually first extract text pixels from the segmentation map and then obtain bounding boxes of the text by post-processing. Zhang \textit{et al.} \cite{ZhangMutidetectFCN2016} used the FCN and MSER for pixel-level multi-oriented text detection. Mask textspotter \cite{Lyumasktextspotter2018} built a model based on the framework of Mask R-CNN \cite{he_maskRCNN_2017} and performed character-level instance segmentation for each alphabet. TextSnake \cite{long_textsnake_2018} proposed a novel representation of arbitrarily shaped text and predicted heat maps of text center lines, text regions, radii, and orientations to extract text regions. PSENet \cite{LiPSENet2020} gradually expanded the text region from small to large kernels to make final predictions through multiple semantic segmentation maps. Liao \textit{et al.} \cite{LiaoDB2020} proposed a module called differentiable binarization (DB) to perform the binarization process in a segmentation network. CRAFT \cite{BaekCRAFT2019} proposed learning each character center and the affinity between characters in the form of a heat map.

\subsection{Image Inpainting} % 2.2
Image inpainting fills the hole regions of an image with plausible content. Image inpainting research can generally be divided into two categories: non-learning and deep-learning-based approaches. Non-learning approaches transfer the surrounding content to the hole region based on low-level features using a traditional algorithm such as patch matching \cite{efros_image_2001,barnes_patchmatch_2009,darabi_inpaintpatch_2012} and diffusion \cite{Bertalmio2000,Oliveira2001}. Although these methods can work well on small holes, they cannot deal with large missing regions, which require semantics patches for the hole restoration based on a high-level understanding of the whole image. 

The generative adversarial network (GAN) strategy has been widely adopted in recent deep-learning-based approaches and can be categorized as single-stage inpainting and progressive inpainting. Context encoders \cite{Pathakcontextencoder2016} first trained an image-inpainting deep neural network using an encoder--decoder structure and adversarial losses. Iizuka \textit{et al.} \cite{iizuka_globally_2017} adopted dilated convolution and proposed global and local discriminators for adversarial training. Liu \textit{et al.} \cite{liu_partialConv_2018} defined a partial convolutional layer with a mask-update mechanism to ensure that the partial convolution filters learn more valid information from the non-hole region and can robustly handle holes of any shape. Xie \textit{et al.} \cite{xie_LBAM_2019} proposed a learnable bidirectional attention module that included forward and reverse attention maps for more effective hole filling. Li \textit{et al.} \cite{li_recurrent_2020} exploited the correlation between adjacent pixels by recurrently inferring and gathering the hole boundary for the encoder, referred to as the recurrent feature reasoning module.

To generate a more realistic texture, a coarse-to-fine strategy was adopted in the progressive inpainting methods. Yu \textit{et al.} \cite{Yu_ContextualAttention_2018} exploited textural similarities to borrow feature information from known background patches to generate missing patches and designed a two-stage network architecture from coarse to fine. Yu \textit{et al.} \cite{yu_gatedConv_2019} introduced gated convolution, which further generalizes partial convolution to make the mask-update mechanism learnable. The method was then combined with the SN-PatchGAN discriminator to obtain better performance. Yi \textit{et al.} \cite{yi_HiFill_2020} improved gated convolution through a lightweight design and proposed high-frequency residuals to generate rich and detailed textures for efficient ultra-high-resolution image inpainting.

\subsection{Text Erasing} % 2.3
Early text erasing research starts with the removal of born-digital text such as watermarks, captions, and subtitles on images or video sequences \cite{Leetextremovalvideo2003}. Owing to the plain layout, color, and regular font, born-digital text can be detected by traditional feature engineering approaches such as binarization \cite{Pnevmatikakis2008,mosleh_texterasing_2012}, and inpainted by patch matching \cite{Khodadadi2012} or smoothing algorithms \cite{wagh_textremoval_2015}.

Erasing text in the wild is a more complex and challenging task because of the various fonts with different layouts and illumination conditions. The recent rapid development of deep neural networks has made scene text erasure a promising research task in the computer vision field. Scene text erasing research can be classified into two categories: one-step and two-step methods. One-step methods use an end-to-end model to directly output a text-erased image without the aid of text location information. Pix2Pix \cite{Isolapixel2pixel2017} is a conditional generative adversarial network (cGAN) designed for general-purpose image-to-image translation tasks that can be applied to text removal. Nakamura \textit{et al.} \cite{Nakamura2017} made the first attempt to build a one-stage scene text eraser, a sliding window method using a skip-connected auto-encoder. This method destroys the integrity of text strokes but cannot maintain the global consistency of the entire image. EnsNet \cite{zhang_Ensnet_2019} adopted cGAN with a refined loss function and a local-aware discriminator to erase texts at the entire image level. Liu \textit{et al.} \cite{LiuEraseNet2020} provided a comprehensive real-world scene-text removal benchmark, named SCUT-EnsText, and proposed EraseNet, which adopts a coarse-to-fine erasure network structure with a segmentation module to generate a mask of the text region to help with text region localization. MTRNet++ \cite{TursunMTRNetpp2020} shared the same coarse-to-fine inpainting idea but used a multi-branch generator. The mask-refine branch predicts stroke-level text masks to guide text removal.

Two-step methods remove the text with an awareness of text location, which can be provided by users or by a pretrained text detector. Qin \textit{et al.} \cite{QinBMVC2018} proposed a cGAN with one encoder and two decoder architectures to perform content region segmentation and inpainting jointly, which can be applied for text removal in cropped text images. MTRNet \cite{TursunMTRNet2019} used manually provided text masks to guide the training and prediction processes of cGAN, thereby realizing the controllability of the text-erasing region. Zdenek \textit{et al.} \cite{Zdenekweaksupervision2020} proposed a weak supervision method employing a pretrained scene text detector \cite{LiPSENet2020} and a pretrained image inpainting model \cite{ZhengPIC2019} to free the scene text erasing task from the requirement of paired-wise training data of scene images with text and the corresponding text-erased images. Bian \textit{et al.} \cite{Biancascaded2020} proposed a cascaded GAN-based model to decouple text stroke detection and stroke removal in text removal tasks.

We consider our proposed method to be a practical solution to scene text erasing tasks because the pretrained scene text detector and synthetic data obviate the need for paired real-world data. The relatively weak text detection ability of one-step end-to-end methods can lead to excessive erasure of text-free areas and incomplete erasure of the text region. The detection ability of these networks can only be improved during end-to-end training, which would require expensive real-world training data. The advantage of our method over weak supervision methods is that using improved synthetic data for training can greatly reduce the domain shift problem during the inpainting process.

\section{Methodology} %3 Methodology
\label{Methodology}
The pipeline of the proposed method is shown in Fig.~\ref{tang2}. Given the source image containing the scene text and the corresponding text bounding boxes, text images are cropped from the source image. Subsequently, each cropped text image goes through the text-erasing network. The cropped text-erased images are then inserted into the source image to obtain an output image that does not contain text.

Our network comprises 1) a stroke mask prediction module and 2) a background inpainting module, as illustrated in Fig.~\ref{tang3}. Specifically, the stroke mask prediction module first predicts the stroke mask of the scene text $\hat{I}_{mask}$ as the hole from $I_{in}$. Then, the background inpainting module is used to fill the hole of the input image $I_{in}$ with the appropriate content and output text-erased image $\hat{I}_{out}$.

\subsection{Stroke Mask Prediction Module} %3.1
% \label{ssec:subhead}
An encoder--decoder FCN is adopted in this module. The input image $I_{in}$ is encoded by three down-sampling convolutional layers and four residual blocks \cite{HeResNet2016}, and the feature maps $F_{m}$ are decoded by three up-sampling transposed convolutional layers to generate the text stroke mask $\hat{I}_{mask}$. The architecture of the SMPM is presented in Table~\ref{tab3.1}. The skip connections concatenate feature maps of the same shape between the encoder and decoder. The feature maps $F_{m}$ output from the residual blocks are concatenated with the feature maps $F_{b}$ in the background inpainting module, which is introduced in Section B. There is usually an imbalance between the pixels of the text region and the pixels of the non-text region in $I_{in}$, and both high recall and precision are expected during mask prediction. Therefore, the dice loss \cite{Milletariunet2016} and L1 loss are employed to guide the generation of the text mask. In image segmentation, the dice loss is a region-based loss measure that expresses the proportion of correctly predicted pixels to the sum of the total pixels of both the prediction and ground truth. Mathematically, the dice loss is defined as
\begin{equation}
\label{L_dice}
\mathcal{L}_{dice} = 1 - \frac{2 \sum_{i}^N (\hat{I}_{mask})_{i} (I_{mask})_{i} }{\sum_{i}^N (\hat{I}_{mask})_{i} + \sum_{i}^N (I_{mask})_{i}},
\end{equation}
where N denotes the total number of pixels in the input image, and $\hat{I}_{mask}$ and $I_{mask}$ represent the prediction and ground truth of the text mask, respectively. The total loss of the stroke mask prediction module is 
\begin{equation}
\label{L_SMPM}
\mathcal L_{SMPM} =  \|\hat{I}_{mask} - I_{mask}\|_{1}  + \lambda_{0} \mathcal L_{dice},
\end{equation}
and in our experiments, $\lambda_{0}$ is set to 1.0.

\begin{table}[]
\centering
\caption{Architecture of the SMPM. Conv: Convolutional Layer, Deconv: Transposed Convolutional Layer, ResBlock: Residual Convolutional Block, C: Channels, K: Kernel Size, S: Stride}
\setlength{\tabcolsep}{5.5mm}{
\label{tab3.1}
\begin{tabular}{ccc}
% \hline
\toprule
Layers     & Conﬁgurations        & Output  \\ \hline
Conv×2     & c: 32, k: 3, s: 1    & 128×640 \\ %\hline
Conv       & c: 64, k: 3, s: 2    & 64×320  \\ \hline
Conv×2     & c: 64, k: 3, s: 1    & 64×320  \\ %\hline
Conv       & c: 128, k: 3, s: 2   & 32×160  \\ \hline
Conv×2     & c: 128, k: 3, s: 1   & 32×160  \\ %\hline
Conv       & c: 256, k: 3, s: 2   & 16×80   \\ %\hline
Conv×2     & c: 256, k: 3, s: 1   & 16×80   \\ \hline
ResBlock×4 & c: 256, k: 3         & 16×80   \\ \hline
Conv×2     & c: 256, k: 3, s: 1   & 16×80   \\ %hline
Deconv     & c: 128, k: 3, s: 1/2 & 32×160  \\ %\hline
Conv×2     & c: 128, k: 3, s: 1   & 32×160  \\ \hline
Deconv     & c: 64, k: 3, s: 1/2  & 64×320  \\ %\hline
Conv×2     & c: 64, k: 3, s: 1    & 64×320  \\ \hline
Deconv     & c: 32, k: 3, s: 1/2  & 128×640 \\ %\hline
Conv×2     & c: 32, k: 3, s: 1    & 128×640 \\ %hline
Conv       & c: 3, k: 3, s: 1     & 128×640 \\ %\hline
\bottomrule
\end{tabular}}
\end{table}

\begin{table}[]
\centering
\caption{Architecture of the BIPM. PConv: Partial Convolutional Layer, Upsample: Upsample Layer, Scale: Scale Factor, C: Channels, K: Kernel Size, S: Stride}
\setlength{\tabcolsep}{6mm}{
\label{tab3.2}
\begin{tabular}{ccc}
\toprule
Layers   & Conﬁgurations      & Output  \\ \hline
PConv    & c: 64, k: 7, s: 2  & 64×320  \\ %\hline
PConv×2  & c: 64, k: 3, s: 1  & 64×320  \\ \hline
PConv    & c: 128, k: 5, s: 2 & 32×160  \\ %\hline
PConv×2  & c: 128, k: 3, s: 1 & 32×160  \\ \hline
PConv    & c: 256, k: 3, s: 2 & 16×80   \\ %\hline
PConv×2  & c: 256, k: 3, s: 1 & 16×80   \\ \hline
GCblock  & c: 512, ratio: 4   & 16×80   \\ %\hline
Conv     & c: 256, k: 3, s: 1  & 16×80   \\ \hline
Upsample & scale: 2            & 32×160  \\ %\hline
Pconv×2  & c: 256, k: 3, s: 1  & 32×160  \\ %\hline
PConv    & c: 128, k: 3, s: 1 & 32×160  \\ \hline
Upsample & scale: 2            & 64×320  \\ %\hline
Pconv×2  & c: 128, k: 3, s: 1 & 64×320  \\ %\hline
PConv    & c: 64, k: 3, s: 1  & 64×320  \\ \hline
Upsample & scale: 2            & 128×640 \\ %\hline
Pconv×2  & c: 64, k: 3, s: 1  & 128×640 \\ %\hline
PConv    & c: 3, k: 3, s: 1   & 128×640 \\ %\hline
\bottomrule
\end{tabular}}
\end{table}

% gc blcok
\begin{figure}[!t]
\includegraphics[width=\linewidth]{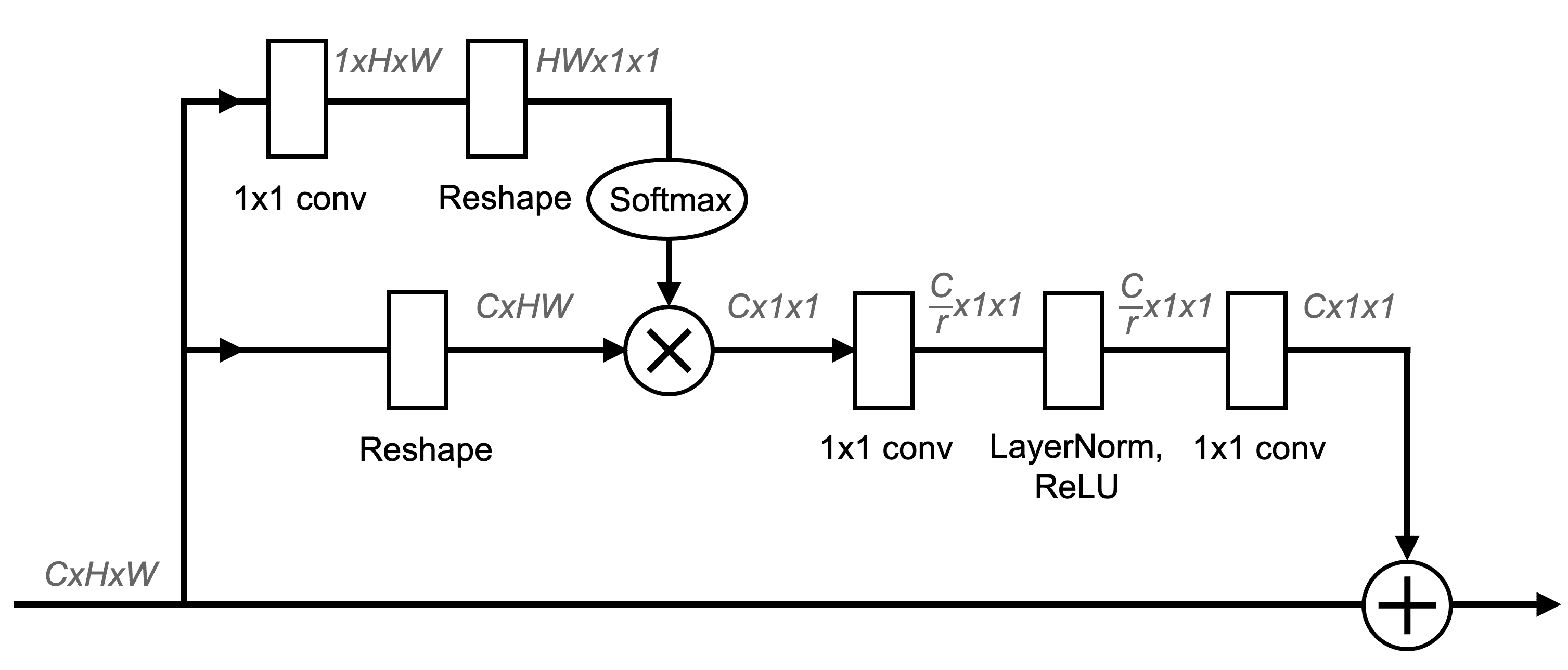}
\caption{Architecture of self-attention block: Global Context (GC) block.} \label{tang4}
\end{figure}

\subsection{Background Inpainting Module} % 3.2
In this module, the input image $I_{in}$ and the predicted mask image $\hat{I}_{mask}$ are taken as input, and the output is the background image $O_{b}$, in which all text stroke pixels, including text shadows caused by illumination, are replaced with proper texture. As shown in the blue part of Fig.~\ref{tang3}, the input image is encoded by three down-sampling partial convolutional layers and concatenated with the feature map $F_{m}$ from the SMPM. This is followed by a self-attention block, which allows the network to learn long-range dependencies. Finally, the decoder generates the output image $\hat{I}_{out}$. The architecture of the BIPM is presented in Table~\ref{tab3.2}.

\subsubsection{Partial Convolutional Layers} % 3.2.1
To generate clearer background images and suppress text ghosts and artifacts, we use partial convolutional layers \cite{liu_partialConv_2018} to allow the network to learn more features from the non-text part of $I_{in}$. The partial convolution layer comprises two steps: the partial convolution operation and mask update. The partial convolution operation and mask update are defined as follows:
\begin{equation}
\label{partial convolution operation}
{x'}=\left\{
\begin{array}{l}
{\textbf W}^{\textit T}(\textbf X  \odot \textbf M)\frac{\text{ sum($\textbf 1$)}}{\text{sum(\textbf M)}} + b, \quad {\text{if sum(\textbf M)}} > 0\\
0, \qquad\qquad\qquad\qquad\qquad\quad \text{otherwise}
\end{array}
\right.
,
\end{equation}

\begin{equation}
\label{mask update}
{m'}=\left\{
\begin{array}{l}
1, \quad \text{if sum(\textbf M)} > 0\\
0, \quad \text{otherwise}
\end{array}
\right.
,
\end{equation}
where $\textbf W$ and ${b}$ indicate the weights and bias of the convolution filter, respectively. $\textbf X$ are the input pixels for the current convolution window, and $\textbf M$ is the corresponding mask in the receptive field. ${\odot}$ denotes element-wise multiplication. $x'$ is the output feature value of the partial convolution, and $m'$ is the updated mask value. $\textbf 1$ is an all-one matrix with the same shape as $\textbf M$.

\subsubsection{Skip Connection and Self-Attention Block} % 3.2.2
By leveraging partial convolutional layers, we can extract more features from outside the text region, which are beneficial for background reconstruction. However, in some cases, the feature information related to texture and illumination (such as highlight, shadow, and transparency) inside the text region is also helpful for background inpainting. Thus, considering that the feature maps $F_{m}$ generated in the SMPM contain rich feature information of the text, we concatenate the feature maps $F_{m}$ and $F_{b}$ from the two modules using skip connection, and feed the concatenated feature maps into a self-attention network that learns both the correspondences between feature maps and non-local features. Consequently, the features inside and outside the text regions are split and weighted by a self-attention block for later decoding. Here, we adopt a Global Context (GC) block \cite{cao_GCNet_2019} as the self-attention module, the architecture of which is illustrated in Fig.~\ref{tang4}. 
% The concatenated features would be ﬁnally fed into a decoder to generate the stylish text

\subsubsection{Training Loss} % 3.2.3
We introduce four loss functions to measure the structural and textural differences between the output image $\hat{I}_{out}$ and the final ground truth ${I}_{out}$, also considering the spatial smoothness of the inpainted region of $\hat{I}_{out}$ during the training of the background inpainting module, which include pixel reconstruction loss, perceptual loss, style loss, and total variation loss. Details of the loss functions are presented below.

Pixel loss is aimed at guiding pixel-level reconstruction, where more weight is added to the inpainted region. The pixel loss can be formulated as follows:
\begin{equation}
\label{L_pixel} % L_pixel
\begin{aligned}
\mathcal L_{pixel} =& \ \| \hat{I}_{mask} \odot (\hat{I}_{out} - I_{out})\|_{1}  \\
                    &+ 6  \|(1-\hat{I}_{mask}) \odot (\hat{I}_{out} - I_{out})\|_{1}.
\end{aligned}
\end{equation}

Perceptual loss and style loss, which are also known as VGG loss \cite{johnson_perceptual_2016, Gatysstyle2016}, are used to make the generated image more realistic. 
Perceptual loss captures high-level semantics and can be considered as a simulation of human perception on images. Perceptual loss computes the differences (L1 loss) between different levels of feature representations between both $\hat{I}_{out}$ and ${I}_{comp}$ and ${I}_{out}$, and is extracted by the same pretrained VGG network \cite{SimonyanVGG2015}. ${I}_{comp}$ is the composed image, where the hole and non-hole regions are from $\hat{I}_{out}$ and ${I}_{out}$, respectively. Perceptual loss can be defined by Eq.~\ref{L_per}:

\begin{equation}
\label{I_comp}
I_{comp} = \hat{I}_{mask} \odot I_{out} + (1-\hat{I}_{mask} )\odot \hat{I}_{out}.
\end{equation}

\begin{equation}
\label{L_per} % L_perceptual
\begin{aligned}
\mathcal L_{per} =& \ \mathbb{E}[\sum\limits_{i=1}{\|\phi_{i}(\hat{I}_{out}) - \phi_{i}(I_{out}) \|_{1}}  \\
                  &+ \sum\limits_{i=1}{\|\phi_{i}({I_{comp}}) - \phi_{i}(I_{out}) \|_{1}}],
                %   \frac{1}{C_{i}H_{i}W_{i}}
\end{aligned}
\end{equation}
where $\phi_{i}$ is the activation map from relu1$\_$1 to the relu5$\_$1 layer of an ImageNet-pretrained VGG-19 model.
% relu1$\_$1 to the relu5$\_$1
Style loss penalizes the differences between both $\hat{I}_{out}$ and ${I}_{comp}$ and ${I}_{out}$ in the image style, such as color, texture, and pattern. The style loss is defined as follows:
\begin{equation}
\label{L_style} % L_style
\begin{aligned}
\mathcal L_{style} =& \ \mathbb{E}_{i}[\|G_{i}^{\phi}(\hat{I}_{out}) - G_{i}^{\phi}(I_{out})\|_{1} \\
                    &+ \|G_{i}^{\phi}(I_{comp}) - G_{i}^{\phi}(I_{out}) \|_{1}],
\end{aligned}
\end{equation}
here, the Gram matrix $G_{i}^{\phi} = \phi_{i}\phi_{i}^{T}/C_{i}H_{i}W_{i}$ and $C_{i}H_{i}W_{i}$ is the shape of the feature map of $\phi_{i}$.

The total variation loss is employed to maintain spatial continuity and smoothness in the generated image to reduce the effect of noise.
\begin{equation}
\label{L_tv} % L_tv
\begin{aligned}
\mathcal L_{tv} =& \ \sum\limits_{(i,j,j+1)\in M}{\| I_{comp}^{i,j+1} - I_{comp}^{i,j} \|_{1}} \\
                &+ \sum\limits_{(i,j,i+1)\in M}{\| I_{comp}^{i+1,j} - I_{comp}^{i,j} \|_{1}},
\end{aligned}
\end{equation}
where $M$ is the hole region in $\hat{I}_{mask}$.

% figure.5
\begin{figure*}[!t]
\includegraphics[width=\textwidth]{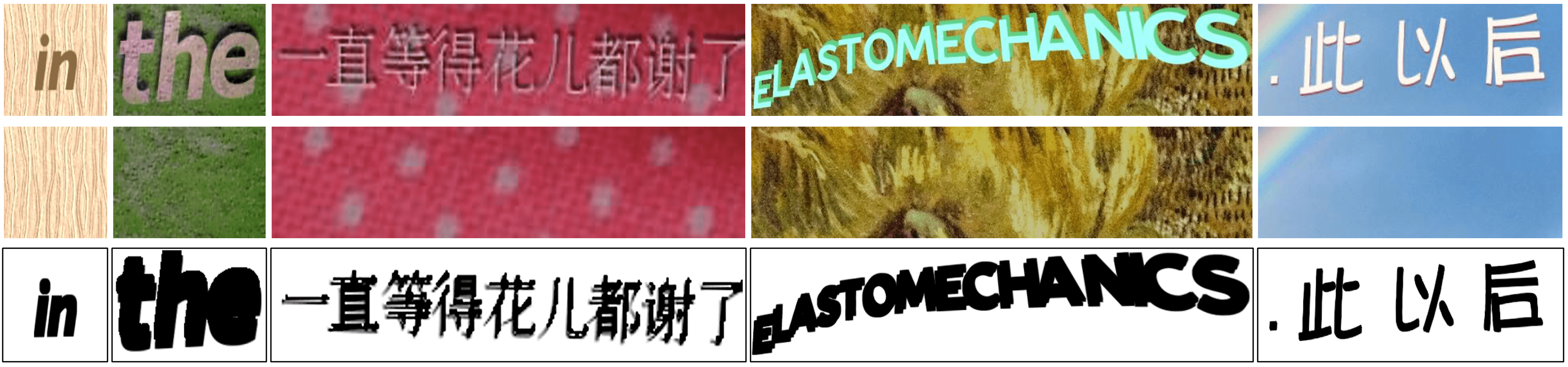}
\caption{Some training image samples generated by our enhanced synthesis text engine. Input (top), final ground truth (middle), text mask ground truth (bottom).}
\label{tang5}
\end{figure*}

Finally, the loss function of the entire network is expressed as follows:
\begin{equation}
\label{L}
\mathcal L = 10 \mathcal L_{SMPM} + \mathcal L_{pixel} + \lambda_{1} \mathcal L_{per} + \lambda_{2} \mathcal L_{style} +\lambda_{3} \mathcal L_{tv},
\end{equation}
where $\lambda_{1}$, $\lambda_{2}$, and $\lambda_{3}$ are set to 0.05, 100, and 0.1, respectively, according to Liu \textit{et al.} \cite{liu_partialConv_2018}. We slightly changed the weight of the style loss according to our own training loss curve.

\section{Experiment} %4
\label{Experiment}
\subsection{Implementation Details} % 4.1
Our implementation is based on PyTorch. In the training process, we generated one million synthetic text images and corresponding text mask images as training data from background images that did not contain text. The input size of our network was 128×640, and the height of the training images was first resized to 128 while maintaining the aspect ratio. Then, if the width of the image was insufficient, the remaining pixels were padded with 0 on the right side of the image; otherwise, it was resized to 640. The training batch size was 8 on a single 1080Ti GPU. We used Adam \cite{KingmaAdam2015} to optimize the entire network with ${\beta}$ = (0.9, 0.999) and set the weight decay to 0. The learning rate started with 0.0002 and decayed to nine-tenth after each epoch in the training phase. The network was trained in an end-to-end manner, and we followed the fine-tuning strategy \cite{liu_partialConv_2018}, which freezes the batch normalization parameters in the encoder of the background inpainting module after approximately 10 epochs. 

In the inference process, we first expanded the text bounding box to include more background information and cropped the expanded text region from the image. Then, we resized and padded the cropped text image following the preprocessing strategy of training and fed it to our proposed network for text erasing. Subsequently, we cut off the padding part of the network output and resized the remaining part back to its original size and ratio. Finally, part of the output, which was inside the original bounding box, was pasted back into the source image. The text in an arbitrary quadrilateral annotation was transformed by perspective transformation, and the text in the curved annotation was transformed by thin-plate-spline into rectangular text images before being fed into our network. The network output was transformed back to its original shape and copied to the original position to obtain the final text-erased image.

\subsection{Dataset and Evaluation Metrics} % 4.2

\subsubsection{Synthetic Dataset} % 4.2.1
\begin{itemize}

\item[\textbullet ] \textbf{Improved Synth-text image} We used over 1,500 English and Chinese fonts and 10,000 background images without text to generate a total of one million images for our model training using our enhanced synthesis text engine, which is improved from Synth-text technology \cite{gupta_synthetic_2016}. The training dataset contains original background images as the final ground truth, synthetic text in background images as input, and mask images of synthetic text as the ground truth of the text mask. 

Compared with the vanilla synth-text method, we made several improvements to make our generated data share more similarity with real-world data. 1) There is a 50\% possibility that the text instance is composed directly on the background instead of being generated by Poisson blending, in which case,  some background information behind the text instance would be provided. 2) Some other effects were added to the text instance, such as Gaussian blur to simulate out-of-focus, text shift, text with a 3D structure, and more shadow parameters to make the shadow of the text more realistic, among others. 3) We compressed and saved the image in JPEG format with different compression qualities to handle images of varying quality. 4) A dilation mask of a text instance, including all of its effects, was generated in the mask image to reduce the effect of JPEG artifacts around the text edge. Some samples of the generated images are shown in Fig.~\ref{tang5}.

\item[\textbullet ] \textbf{SCUT-Syn} \cite{zhang_Ensnet_2019} was created by Synthesis text engine \cite{gupta_synthetic_2016}, which contains 8,000 images for training and 800 images for testing. The background images of this dataset were collected from ICDAR 2013 \cite{KaratzasICDAR2013} and ICDAR MLT-2017 \cite{NayefICDAR2017MLT}, and the text instances in the background images were manually removed. Most test images were from the training set, and the training and testing sets were generated from the same background images, although the synthesized text instances were different. We evaluated our method using only test images. 

\end{itemize}

% figure.6
\begin{figure*}[!t]
\centering
\subfigure[Input]{
\begin{minipage}[t]{0.162\textwidth}
    \includegraphics[height=0.85\textwidth, width=1\textwidth]{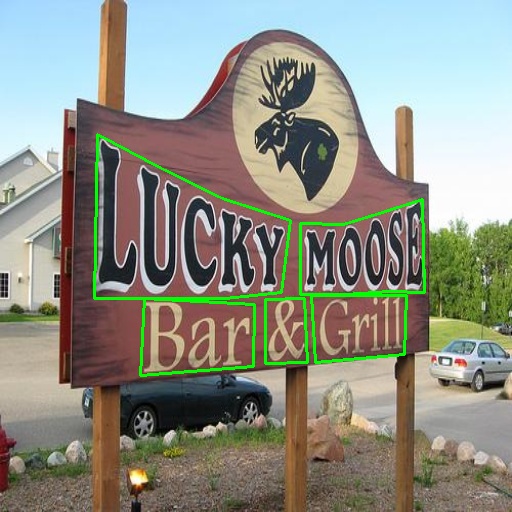}\vspace{1.5pt}
    \includegraphics[height=0.85\textwidth, width=1\textwidth]{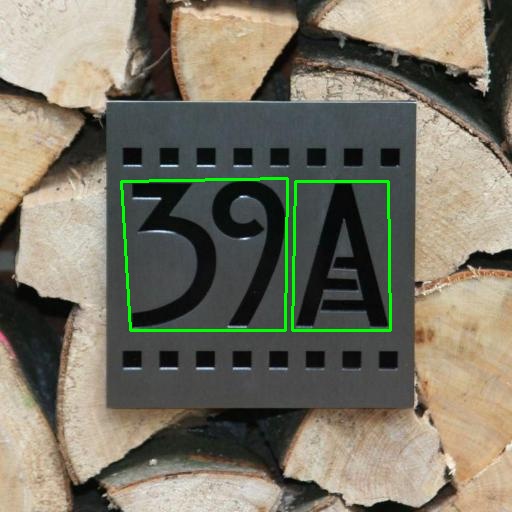}\vspace{1.5pt}
    \includegraphics[height=0.85\textwidth, width=1\textwidth]{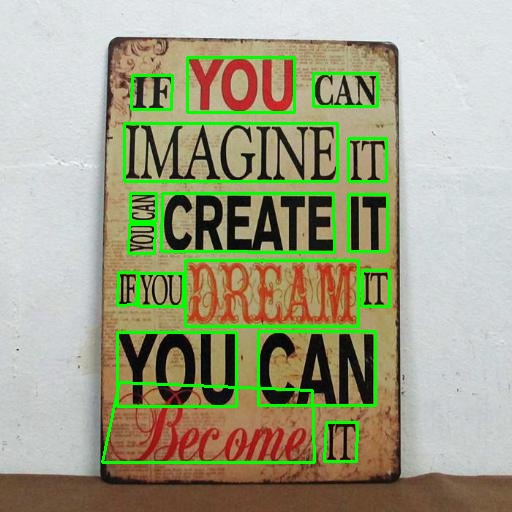}\vspace{4pt}
\end{minipage}}
\hspace*{-9pt}
\subfigure[BIPM]{
\begin{minipage}[t]{0.162\textwidth}
    \includegraphics[height=0.85\textwidth, width=1\textwidth]{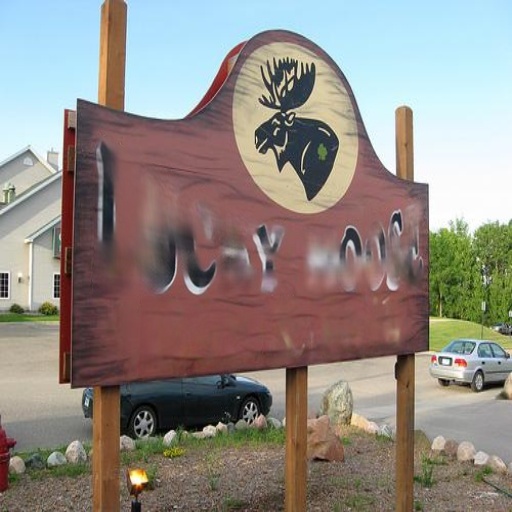}\vspace{1.5pt}
    \includegraphics[height=0.85\textwidth, width=1\textwidth]{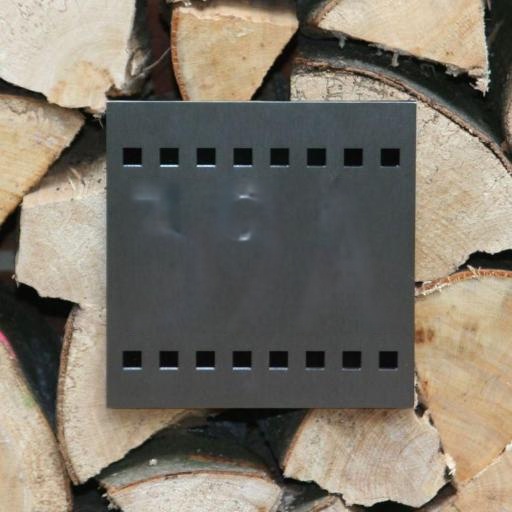}\vspace{1.5pt}
    \includegraphics[height=0.85\textwidth, width=1\textwidth]{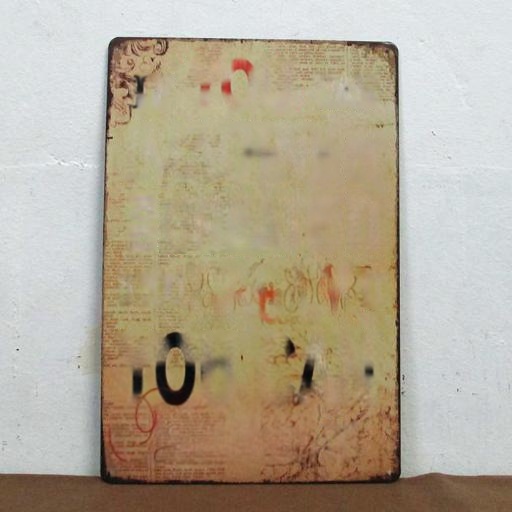}\vspace{4pt}
\end{minipage}}\vspace{1.5pt}
\hspace*{-9pt}
\subfigure[SMPM+BIPM]{
\begin{minipage}[t]{0.162\textwidth}
    \includegraphics[height=0.85\textwidth, width=1\textwidth]{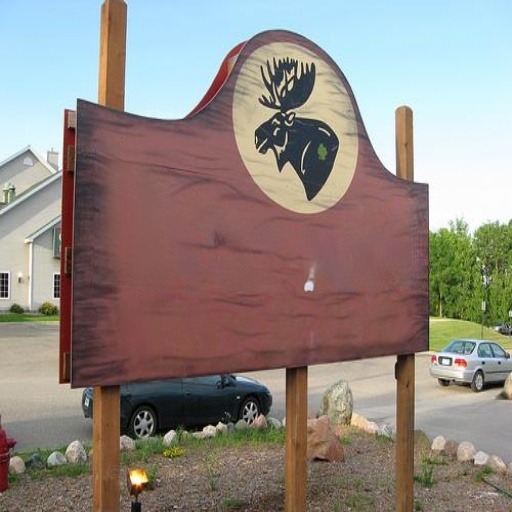}\vspace{1.5pt}
    \includegraphics[height=0.85\textwidth, width=1\textwidth]{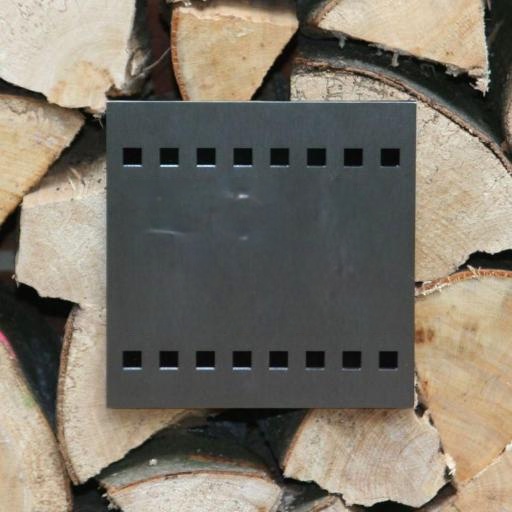}\vspace{1.5pt}
    \includegraphics[height=0.85\textwidth, width=1\textwidth]{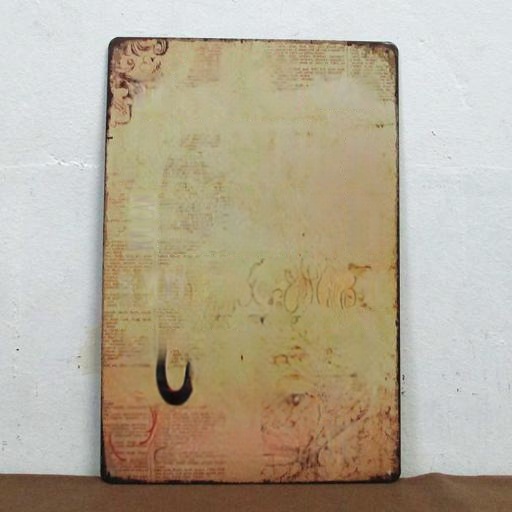}\vspace{4pt}
\end{minipage}}
\hspace*{-9pt}
\subfigure[SMPM+BIPM+SC]{
\begin{minipage}[t]{0.162\textwidth}
    \includegraphics[height=0.85\textwidth, width=1\textwidth]{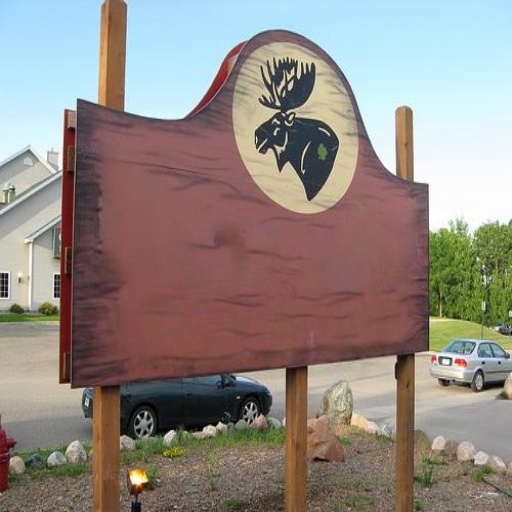}\vspace{1.5pt}
    \includegraphics[height=0.85\textwidth, width=1\textwidth]{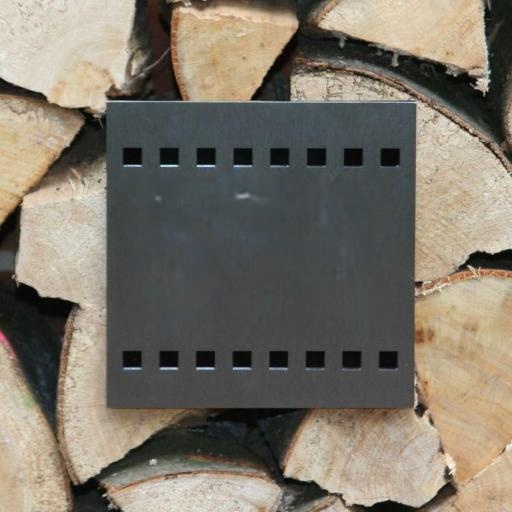}\vspace{1.5pt}
    \includegraphics[height=0.85\textwidth, width=1\textwidth]{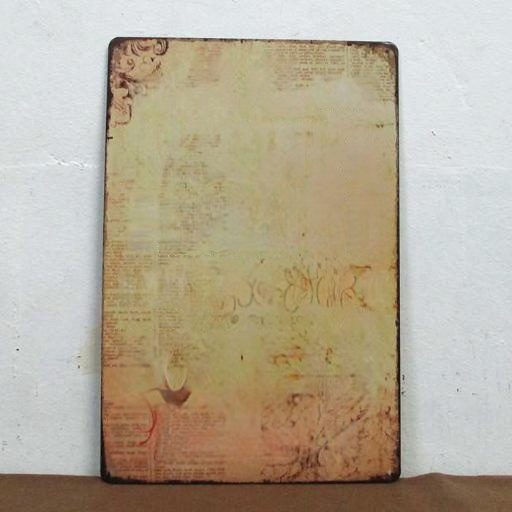}\vspace{4pt}
\end{minipage}}
\hspace*{-9pt}
\subfigure[All (w/o PConv)]{
\begin{minipage}[t]{0.162\textwidth}
    \includegraphics[height=0.85\textwidth, width=1\textwidth]{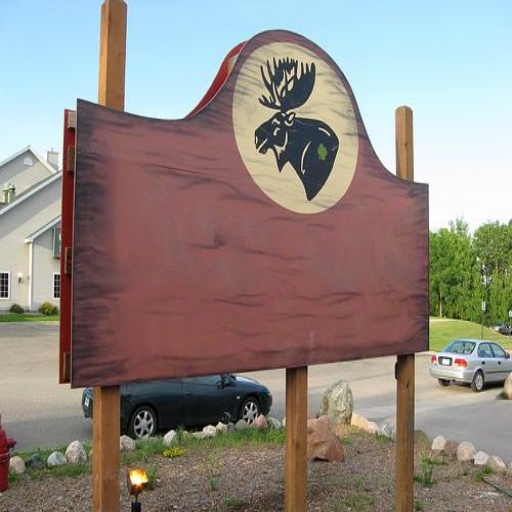}\vspace{1.5pt}
    \includegraphics[height=0.85\textwidth, width=1\textwidth]{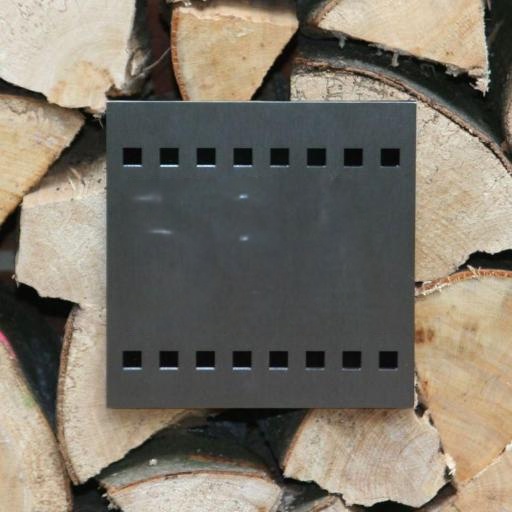}\vspace{1.5pt}
    \includegraphics[height=0.85\textwidth, width=1\textwidth]{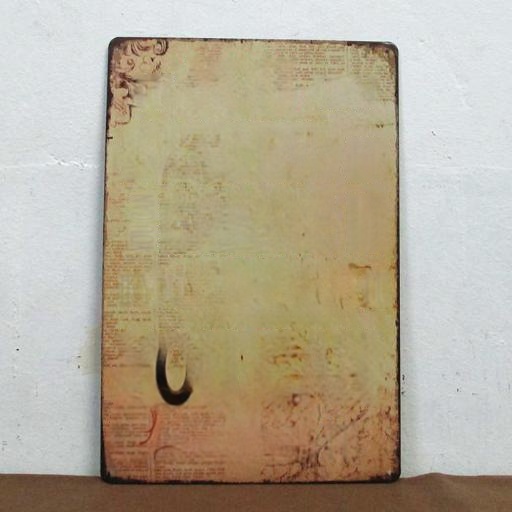}\vspace{4pt}
\end{minipage}}
\hspace*{-9pt}
\subfigure[All]{
\begin{minipage}[t]{0.162\textwidth}
    \includegraphics[height=0.85\textwidth, width=1\textwidth]{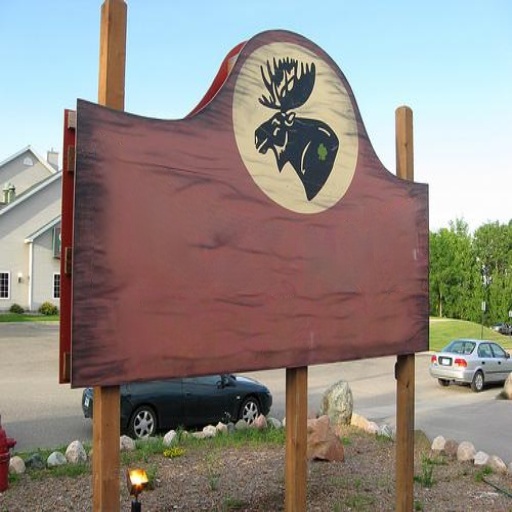}\vspace{1.5pt}
    \includegraphics[height=0.85\textwidth, width=1\textwidth]{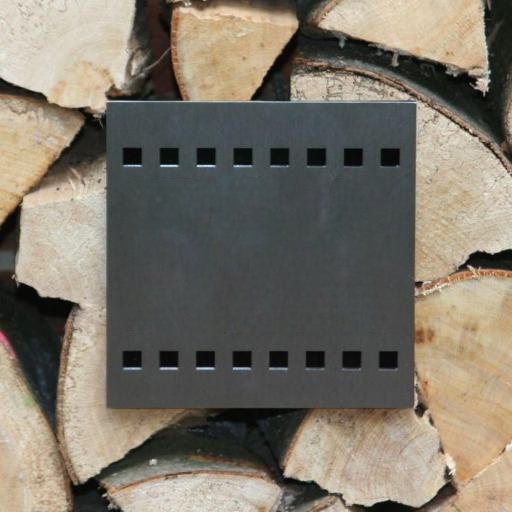}\vspace{1.5pt}
    \includegraphics[height=0.85\textwidth, width=1\textwidth]{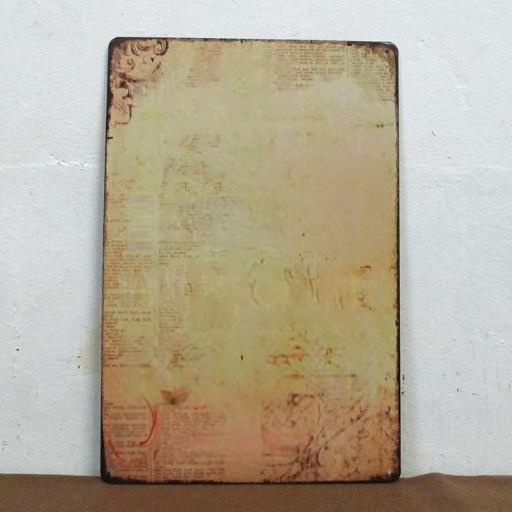}\vspace{4pt}
\end{minipage}}
\vspace{-4pt}
\caption{Visual quality results of ablation study on SCUT-EnsText dataset. From left to right: input images, output of BIPM, output of SMPM+BIPM, output of SMPM+BIPM+SC, output of all (w/o PConv), and output of all. SMPM: Stroke mask prediction module, BIPM: Background inpainting module, PConv: Partial convolutions, SC: Skip connection between two modules, SA: Self-attention block, All: SMPM + BIPM + SC + SA.} \label{tang6}
\end{figure*}

% table.1
\begin{table*}[!t]
\caption{Ablation Study and Qualitative Comparison Between Different Configurations of Our Proposed Network on SCUT-Syn and SCUT-EnsText Datasets. SMPM: Stroke Mask Prediction Module, BIPM: Background Inpainting Module, PConv: Partial Convolutions, SC: Skip Connection between Two Modules, SA: Self-attention Block, Dice: Dice Loss, L1: L1 Loss}
\label{tab4.3}
\centering
\label{tab:my-table}
\begin{tabular}{c|ccc|ccc}
\hline
\multirow{2}{*}{Method}              & \multicolumn{3}{|c}{SCUT-Syn}                                                  & \multicolumn{3}{|c}{SCUT-EnsText}                                              \\ \cline{2-7} 
                                     & \multicolumn{1}{c|}{PSNR↑} & \multicolumn{1}{c|}{SSIM(\%)↑} & MSE↓             & \multicolumn{1}{c|}{PSNR↑} & \multicolumn{1}{c|}{SSIM(\%)↑} & MSE↓             \\ \hline
BIPM (w/o PConv)                     & 34.78                      & 96.67                          & 0.00061          & 34.62                      & 95.79                          & 0.00114          \\
SMPM + BIPM                          & 37.74                      & 97.31                          & 0.00034          & 36.11                      & 96.31                          & 0.00077          \\
SMPM + BIPM + SC                     & 38.29                      & 97.48                          & 0.00030          & 36.36                      & 96.39                          & 0.00070          \\
SMPM + BIPM (w/o PConv) + SC + SA    & \textbf{38.70}             & \textbf{97.67}                 & 0.00025          & 36.49                      & 96.39                          & 0.00073          \\ \hline
SMPM + BIPM + SC + SA w/o Dice & 38.58                      & 97.59                          & 0.00026          & 36.64                      & 96.42                          & 0.00067          \\
SMPM + BIPM + SC + SA w/o L1   & 38.60                      & 97.61                          & 0.00026          & 36.82                      & 96.50                          & 0.00062          \\
SMPM + BIPM + SC + SA (all)          & 38.60                      & 97.55                          & \textbf{0.00024} & \textbf{37.08}             & \textbf{96.54}                 & \textbf{0.00054} \\ \hline
\end{tabular}
\end{table*}

\subsubsection{Real-world Dataset}% 4.2.2
\begin{itemize}

\item[\textbullet] \textbf{ICDAR 2013} \cite{KaratzasICDAR2013} is a widely used scene text images dataset that includes 229 training images and 223 testing images. All text instances were in English and were well-focused. In this study, only the test set was used for the evaluation.

\item[\textbullet] \textbf{SCUT-EnsText} \cite{LiuEraseNet2020} is a comprehensive and challenging scene text removal dataset, containing 2,749 training images and 813 testing images, which are collected from ICDAR2013 \cite{KaratzasICDAR2013}, ICDAR-2015 \cite{KaratzasICDAR2015}, MS COCO-Text \cite{veit_cocotext_2016}, SVT \cite{Wangwordspot2010}, MLT-2019 \cite{NayefICDAR2019MLT}, and ArTs \cite{ChngICDAR2019ArT}. The text instances of this dataset are in Chinese or English with diverse shapes, such as horizontal text, arbitrary quadrilateral text, and curved text. All text instances were carefully erased by annotators with good visual quality. By providing original text annotation and text-erased ground-truth, this dataset can be comprehensively used for both qualitative and quantitative evaluations. In this study, we used this test set to evaluate the performance of the proposed method.

\end{itemize}

\subsubsection{Evaluation metrics}% 4.2.2
\begin{itemize}

\item[\textbullet ] \textbf{Quantitative Evaluation} To quantify the text erasure ability of a model, we followed \cite{TursunMTRNetpp2020,TursunMTRNet2019,zhang_Ensnet_2019,Zdenekweaksupervision2020,LiuEraseNet2020} and utilized a baseline scene text detection model to detect the texts in the text-erased images and evaluated how low are the recall of the detection results. A lower recall indicated that less text was detected, and more text was erased by the model. To make a fair comparison with previous studies, scene text detector EAST and ICDAR 2013 evaluation \cite{zhou_east_2017} protocols were used in the ICDAR 2013 dataset \cite{KaratzasICDAR2013}, text detector CRAFT \cite{BaekCRAFT2019}, and ICDAR 2015 \cite{KaratzasICDAR2015} protocols were adopted in the evaluation of SCUT-EnsText \cite{LiuEraseNet2020}.

\item[\textbullet ] \textbf{Qualitative Evaluation} We followed the previous image inpainting works by reporting metrics including peak signal-to-noise ratio (PSNR), the structural similarity index (SSIM) \cite{WangSSIM2004} and mean squared error (MSE). Higher SSIM, PSNR, and lower MSE values indicate better image restoration quality. Qualitative evaluations were conducted on both the SCUT-Syn \cite{zhang_Ensnet_2019} and SCUT-EnsText \cite{LiuEraseNet2020} datasets.

% \item[\textbullet ] \textbf{Visual quality visual} appearance estimation is also used on real dataset to compare the performance of various methods

\end{itemize}

% figure.7
\begin{figure*}[!ht]
\centering
\subfigure[Input of ours]{
\begin{minipage}[t]{0.241\textwidth}
    \includegraphics[height=0.85\textwidth, width=1\textwidth]{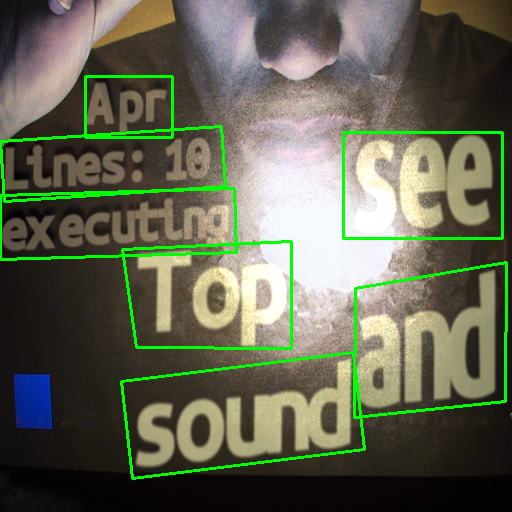}\vspace{2.5pt}
    \includegraphics[height=0.85\textwidth, width=1\textwidth]{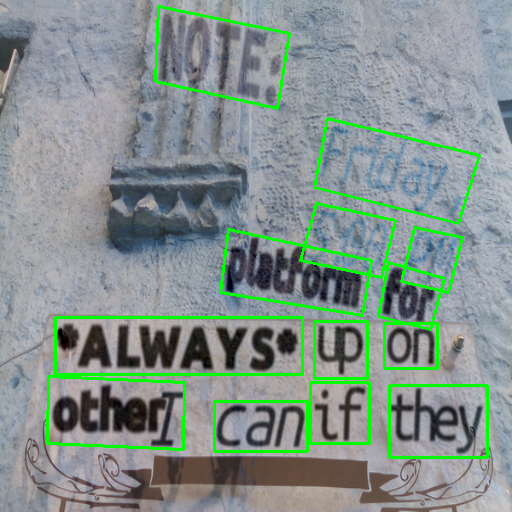}\vspace{2.5pt}
    \includegraphics[height=0.85\textwidth, width=1\textwidth]{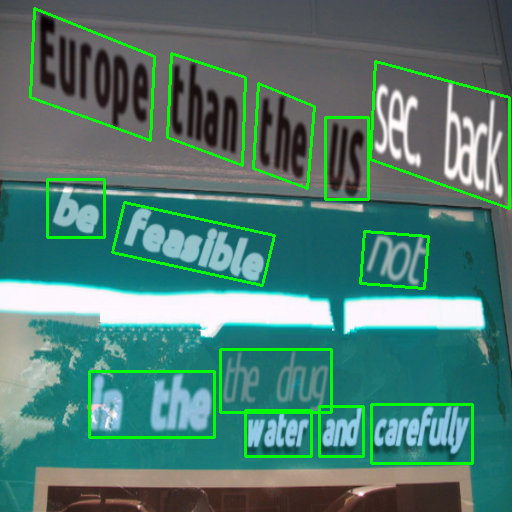}\vspace{4pt}
\end{minipage}}
\hspace*{-8pt}
\subfigure[Ground-truth]{
\begin{minipage}[t]{0.241\textwidth}
    \includegraphics[height=0.85\textwidth, width=1\textwidth]{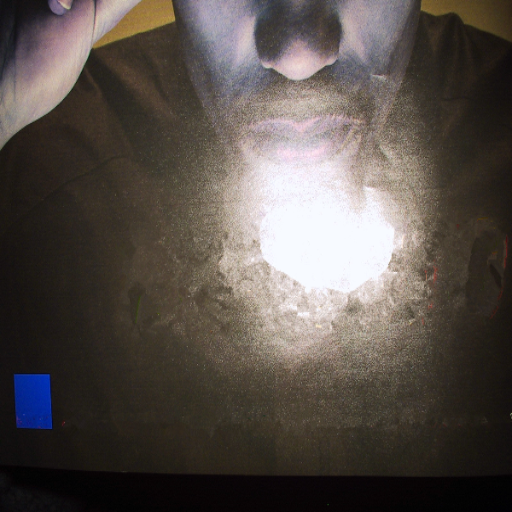}\vspace{2.5pt}
    \includegraphics[height=0.85\textwidth, width=1\textwidth]{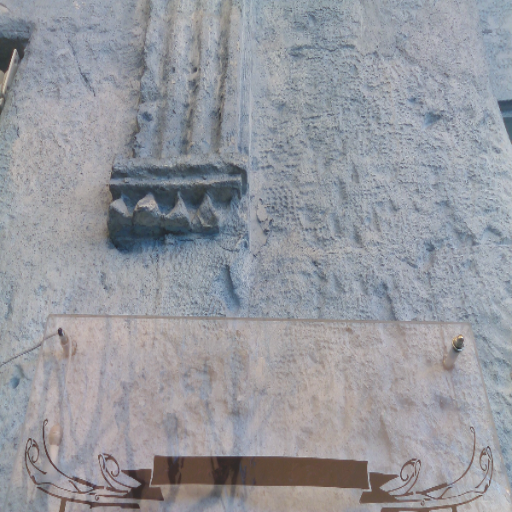}\vspace{2.5pt}
    \includegraphics[height=0.85\textwidth, width=1\textwidth]{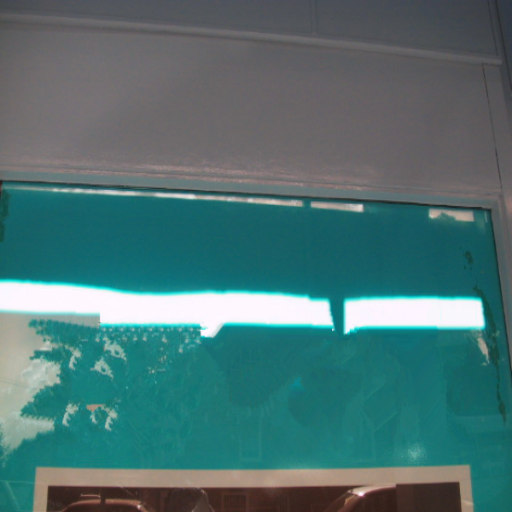}\vspace{4pt}
\end{minipage}}\vspace{2.5pt}
\hspace*{-8pt}
\subfigure[Ours]{
\begin{minipage}[t]{0.241\textwidth}
    \includegraphics[height=0.85\textwidth, width=1\textwidth]{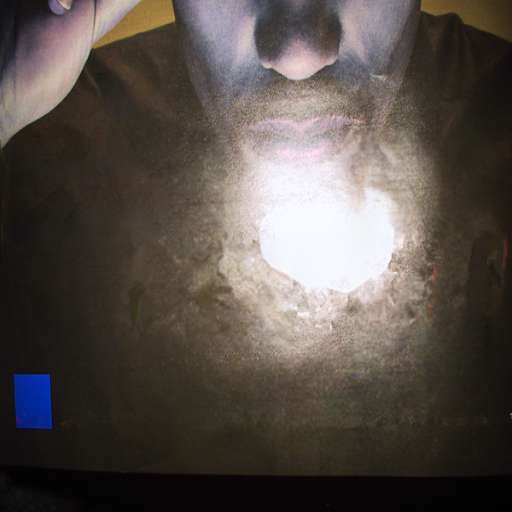}\vspace{2.5pt}
    \includegraphics[height=0.85\textwidth, width=1\textwidth]{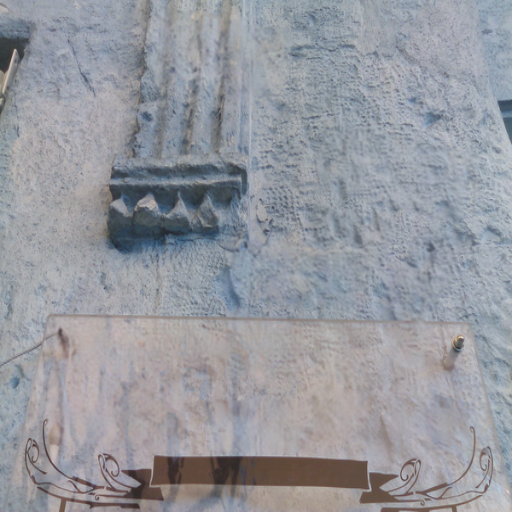}\vspace{2.5pt}
    \includegraphics[height=0.85\textwidth, width=1\textwidth]{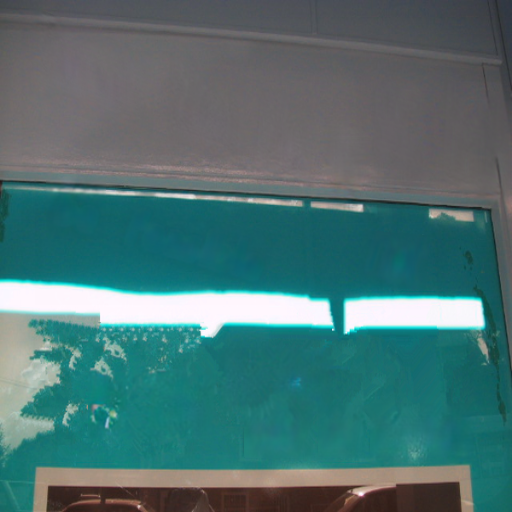}\vspace{4pt}
\end{minipage}}
\hspace*{-8pt}
\subfigure[Predicted text mask]{
\begin{minipage}[t]{0.241\textwidth}
    \setlength{\fboxrule}{0.4pt}
    \setlength{\fboxsep}{0pt}
    \fbox{\includegraphics[height=0.85\textwidth, width=1\textwidth]{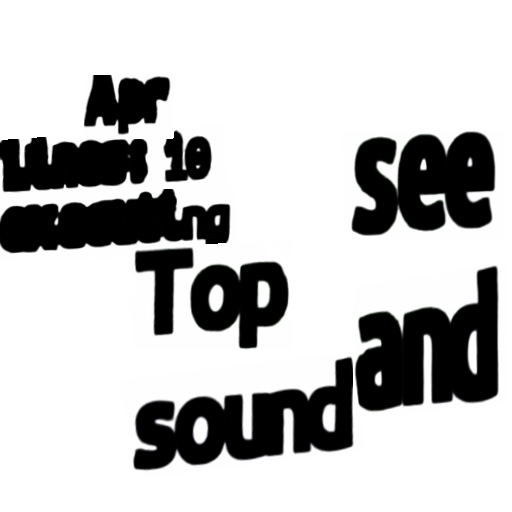}}\vspace{1.7pt}
    \fbox{\includegraphics[height=0.85\textwidth, width=1\textwidth]{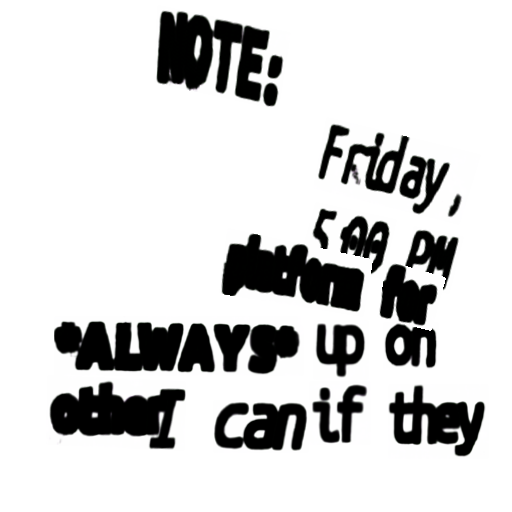}}\vspace{1.7pt}
    \fbox{\includegraphics[height=0.85\textwidth, width=1\textwidth]{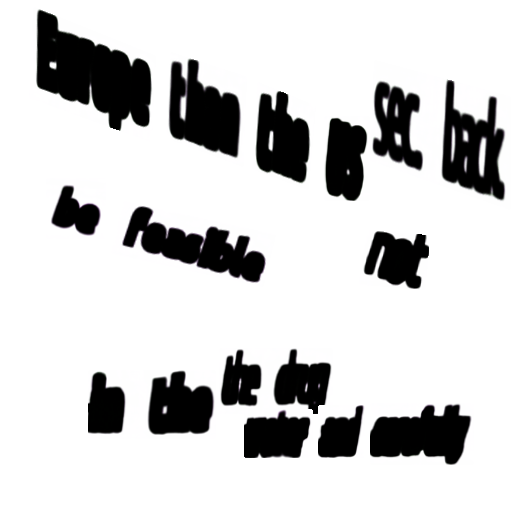}}\vspace{4pt}
\end{minipage}}
\vspace{-4pt}
\caption{Qualitative results of our method on the SCUT-Syn dataset. From left to right: input image and text bounding boxes, ground truth, output of our method, and predicted text mask.} \label{tang7}
\end{figure*}

% figure.8
\begin{figure*}[!t]
\centering
\subfigure[Input of ours]{
\begin{minipage}[t]{0.162\textwidth}
    \includegraphics[height=0.85\textwidth, width=1\textwidth]{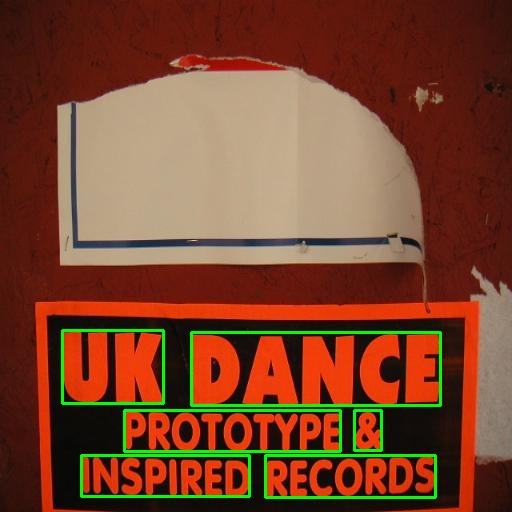}\vspace{1.5pt}
    \includegraphics[height=0.85\textwidth, width=1\textwidth]{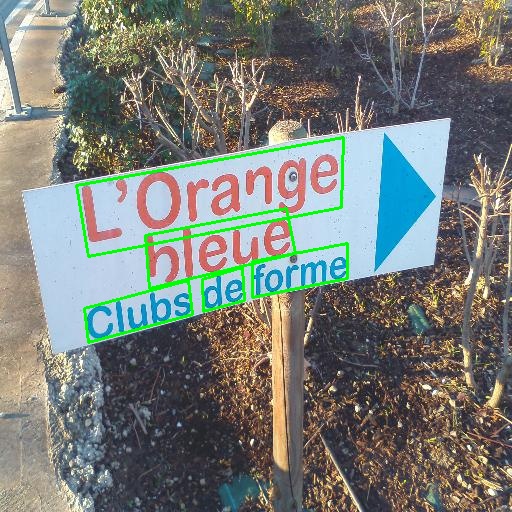}\vspace{1.5pt}
    \includegraphics[height=0.85\textwidth, width=1\textwidth]{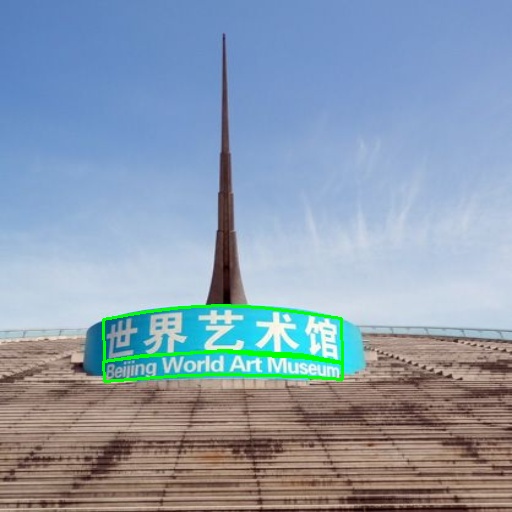}\vspace{1.5pt}
    \includegraphics[height=0.85\textwidth, width=1\textwidth]{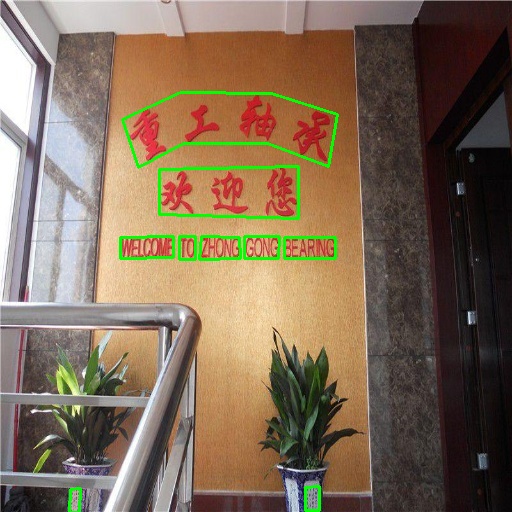}\vspace{4pt}
\end{minipage}}
\hspace*{-9pt}
\subfigure[\fontsize{6.8pt}{6.8pt}\selectfont Input of inpainting methods]{
\begin{minipage}[t]{0.162\textwidth}
    \includegraphics[height=0.85\textwidth, width=1\textwidth]{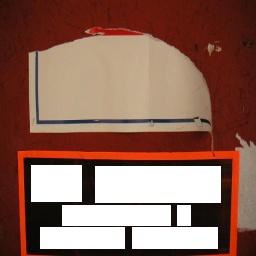}\vspace{1.5pt}
    \includegraphics[height=0.85\textwidth, width=1\textwidth]{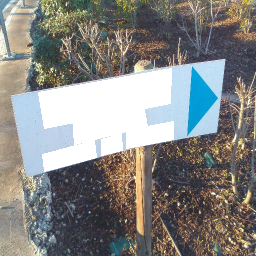}\vspace{1.5pt}
    \includegraphics[height=0.85\textwidth, width=1\textwidth]{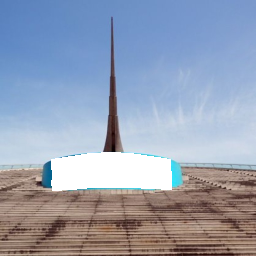}\vspace{1.5pt}
    \includegraphics[height=0.85\textwidth, width=1\textwidth]{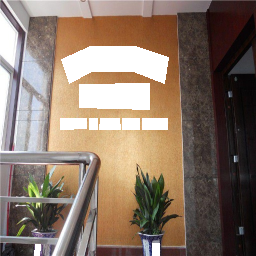}\vspace{4pt}
\end{minipage}}\vspace{1.5pt}
\hspace*{-9pt}
\subfigure[Ours]{
\begin{minipage}[t]{0.162\textwidth}
    \includegraphics[height=0.85\textwidth, width=1\textwidth]{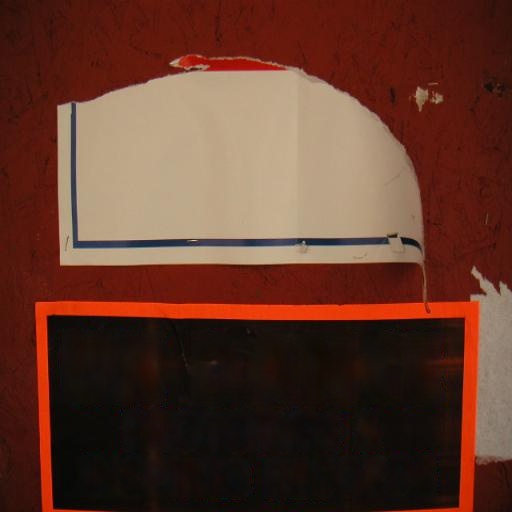}\vspace{1.5pt}
    \includegraphics[height=0.85\textwidth, width=1\textwidth]{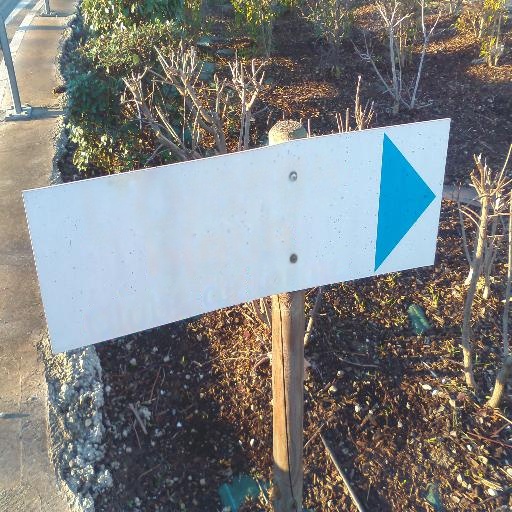}\vspace{1.5pt}
    \includegraphics[height=0.85\textwidth, width=1\textwidth]{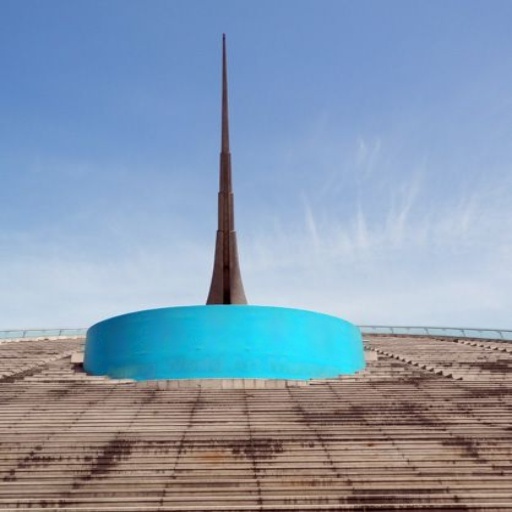}\vspace{1.5pt}
    \includegraphics[height=0.85\textwidth, width=1\textwidth]{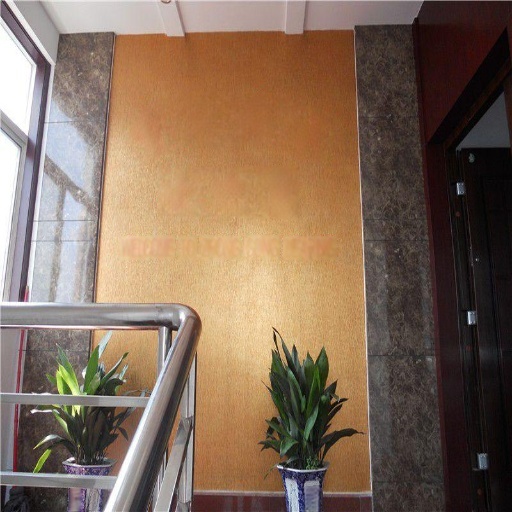}\vspace{4pt}
\end{minipage}}
\hspace*{-9pt}
\subfigure[LBAM]{
\begin{minipage}[t]{0.162\textwidth}
    \includegraphics[height=0.85\textwidth, width=1\textwidth]{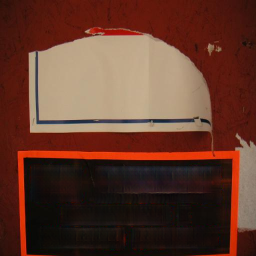}\vspace{1.5pt}
    \includegraphics[height=0.85\textwidth, width=1\textwidth]{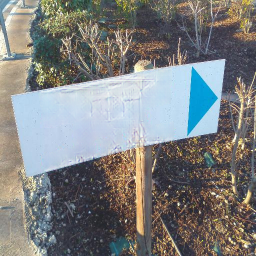}\vspace{1.5pt}
    \includegraphics[height=0.85\textwidth, width=1\textwidth]{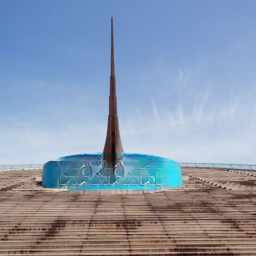}\vspace{1.5pt}
    \includegraphics[height=0.85\textwidth, width=1\textwidth]{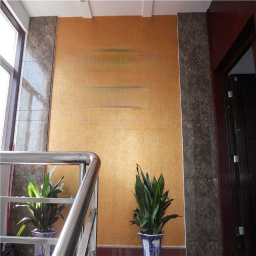}\vspace{4pt}
\end{minipage}}
\hspace*{-9pt}
\subfigure[RFR-Net]{
\begin{minipage}[t]{0.162\textwidth}
    \includegraphics[height=0.85\textwidth, width=1\textwidth]{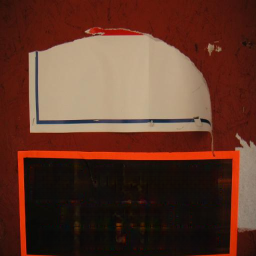}\vspace{1.5pt}
    \includegraphics[height=0.85\textwidth, width=1\textwidth]{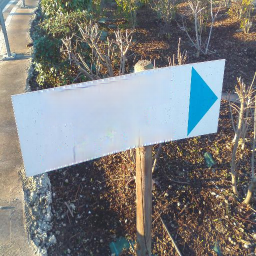}\vspace{1.5pt}
    \includegraphics[height=0.85\textwidth, width=1\textwidth]{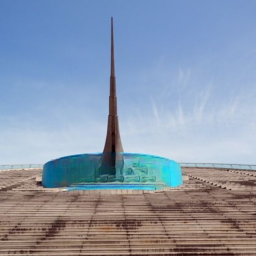}\vspace{1.5pt}
    \includegraphics[height=0.85\textwidth, width=1\textwidth]{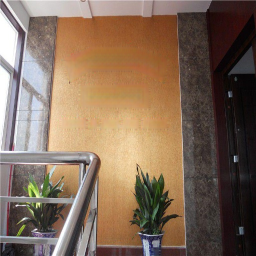}\vspace{4pt}
\end{minipage}}
\hspace*{-9pt}
\subfigure[HiFill]{
\begin{minipage}[t]{0.162\textwidth}
    \includegraphics[height=0.85\textwidth, width=1\textwidth]{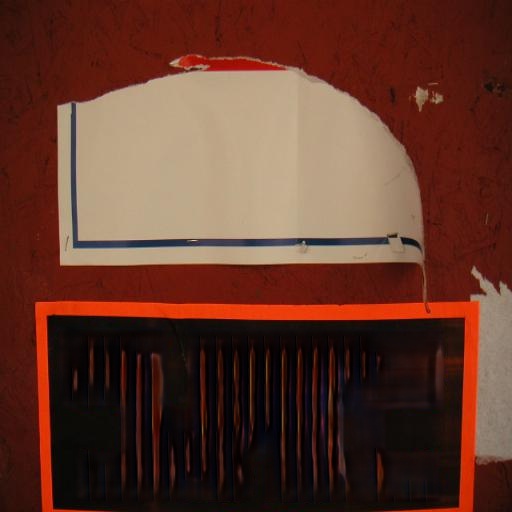}\vspace{1.5pt}
    \includegraphics[height=0.85\textwidth, width=1\textwidth]{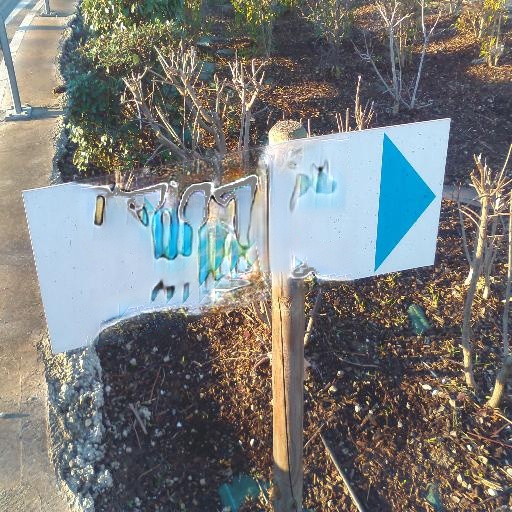}\vspace{1.5pt}
    \includegraphics[height=0.85\textwidth, width=1\textwidth]{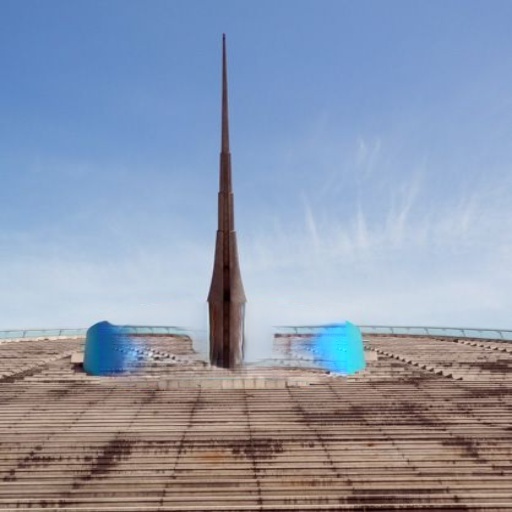}\vspace{1.5pt}
    \includegraphics[height=0.85\textwidth, width=1\textwidth]{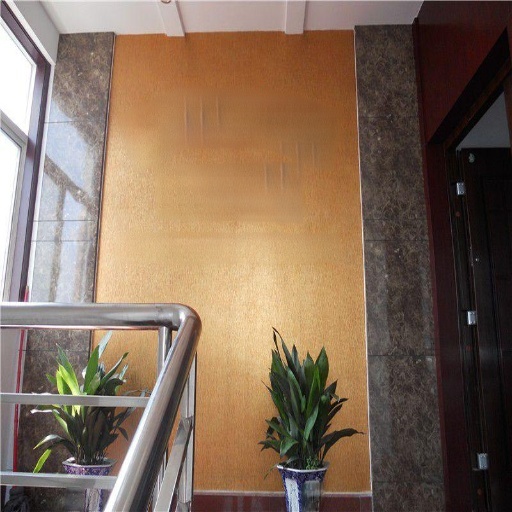}\vspace{4pt}
\end{minipage}}
\vspace{-4pt}
\caption{Visual qualitative comparison between our method and state-of-the-art image inpainting methods on the SCUT-EnsText dataset. From left to right: input of our method, input of inpainting methods, output of our method, output of LBAM, output of RFR-Net, and output of HiFiII.} \label{tang9}
\end{figure*}

% %
% % figure.8
% \begin{figure*}[!t]
% \begin{minipage}{\linewidth}
% \centerline{\includegraphics[width=\textwidth]{tang9.png}}
% \leftline{\qquad Input of ours \ Input of inpainting methods \quad Ours \qquad\qquad\quad LBAM \qquad\qquad\quad RFR-Net \qquad\qquad\quad HiFill}
% \end{minipage}
% \caption{Visual qualitative comparison between our method and state-of-the-art image inpainting methods on the SCUT-EnsText dataset. From left to right: input of our method, input of inpainting methods, output of our method, output of LBAM, output of RFR-Net, and output of HiFiII.} \label{tang9}
% \end{figure*}
% %

% figure.9
\begin{figure*}[!ht]
\centering
\subfigure[Input]{
\begin{minipage}[t]{0.241\textwidth}
    \includegraphics[height=0.85\textwidth, width=1\textwidth]{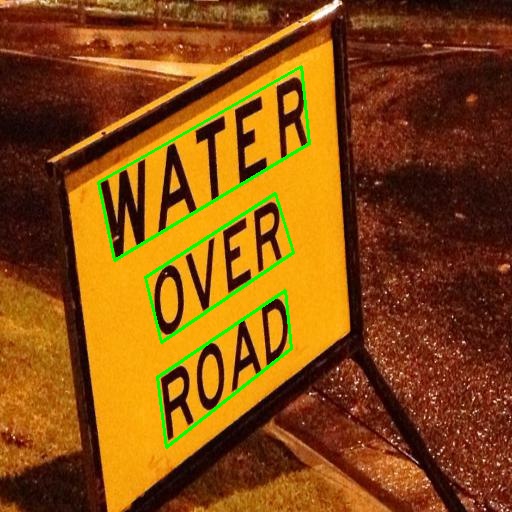}\vspace{2.5pt}
    \includegraphics[height=0.85\textwidth, width=1\textwidth]{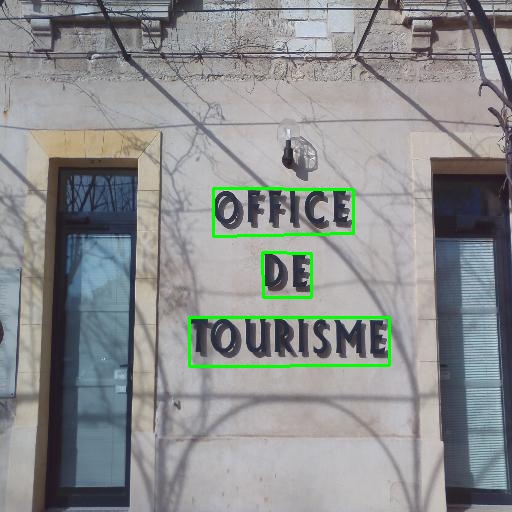}\vspace{2.5pt}
    \includegraphics[height=0.85\textwidth, width=1\textwidth]{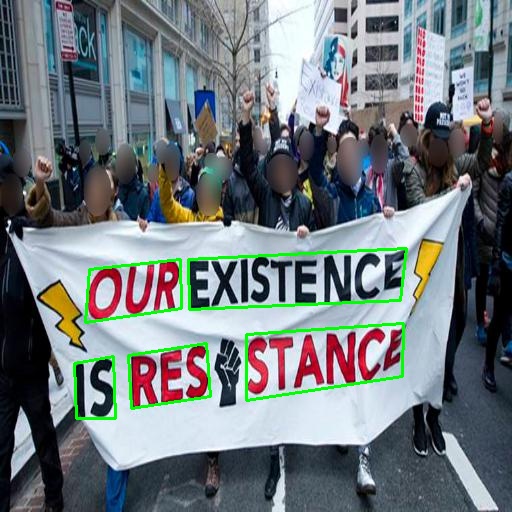}\vspace{2.5pt}
    \includegraphics[height=0.85\textwidth, width=1\textwidth]{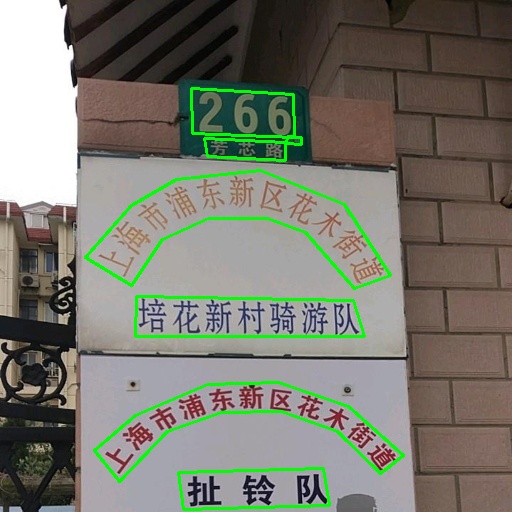}\vspace{2.5pt}
    \includegraphics[height=0.85\textwidth, width=1\textwidth]{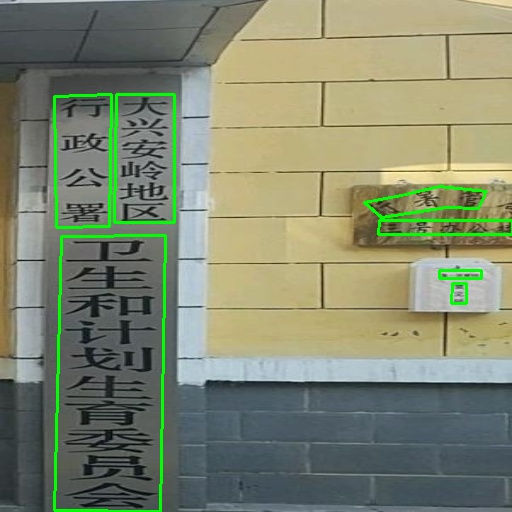}\vspace{2.5pt}
    \includegraphics[height=0.85\textwidth, width=1\textwidth]{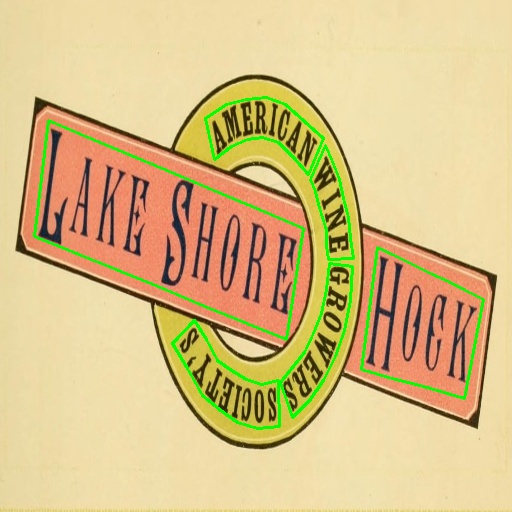}\vspace{4pt}
\end{minipage}}
\hspace*{-8pt}
\subfigure[Ground-truth]{
\begin{minipage}[t]{0.241\textwidth}
    \includegraphics[height=0.85\textwidth, width=1\textwidth]{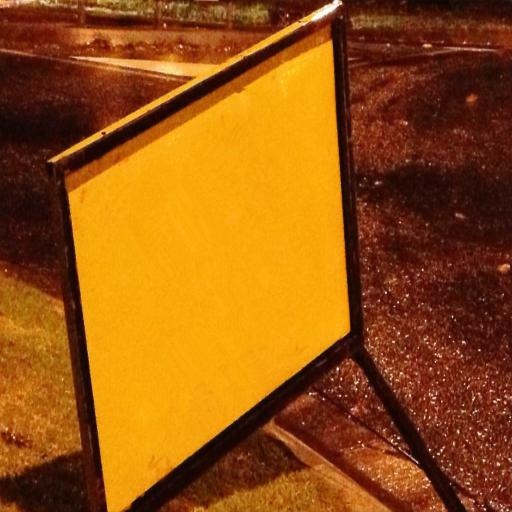}\vspace{2.5pt}
    \includegraphics[height=0.85\textwidth, width=1\textwidth]{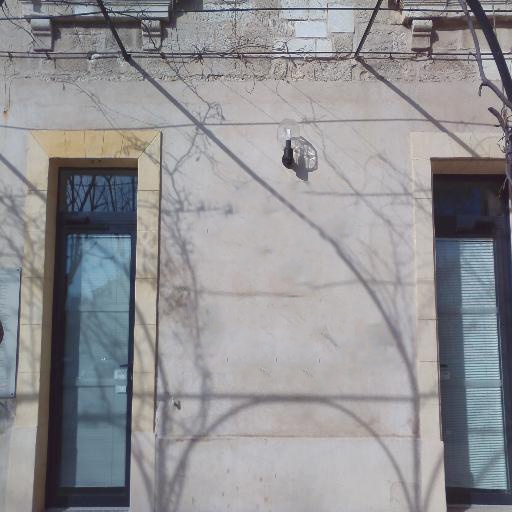}\vspace{2.5pt}
    \includegraphics[height=0.85\textwidth, width=1\textwidth]{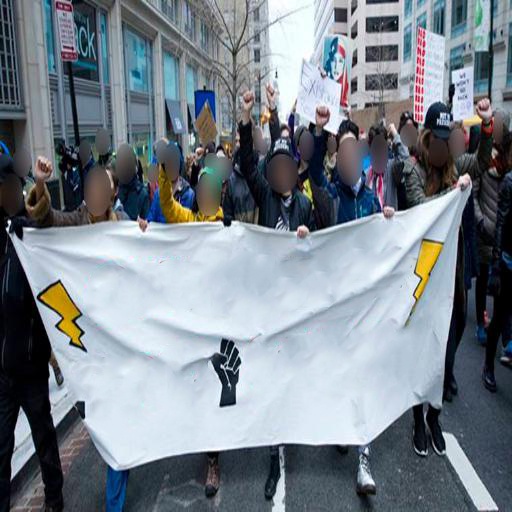}\vspace{2.5pt}
    \includegraphics[height=0.85\textwidth, width=1\textwidth]{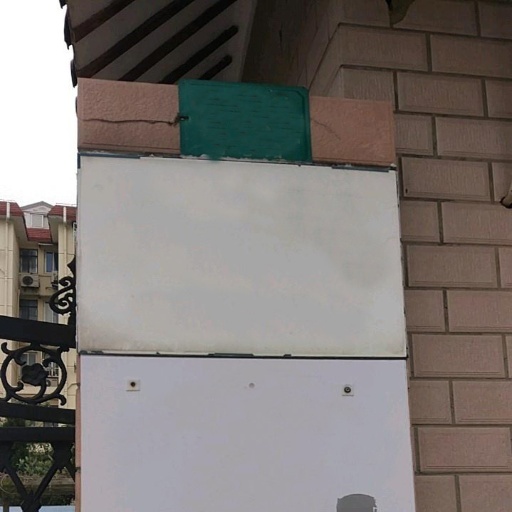}\vspace{2.5pt}
    \includegraphics[height=0.85\textwidth, width=1\textwidth]{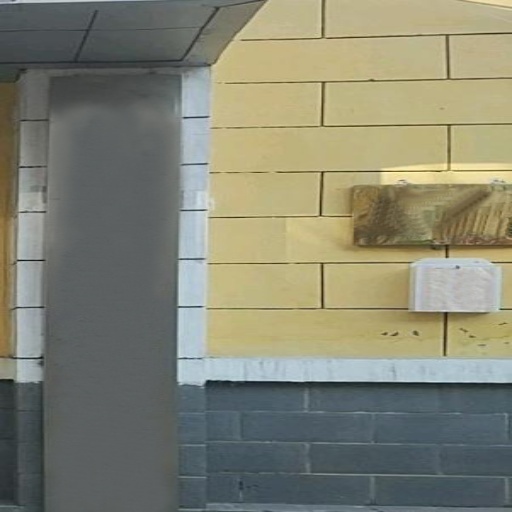}\vspace{2.5pt}
    \includegraphics[height=0.85\textwidth, width=1\textwidth]{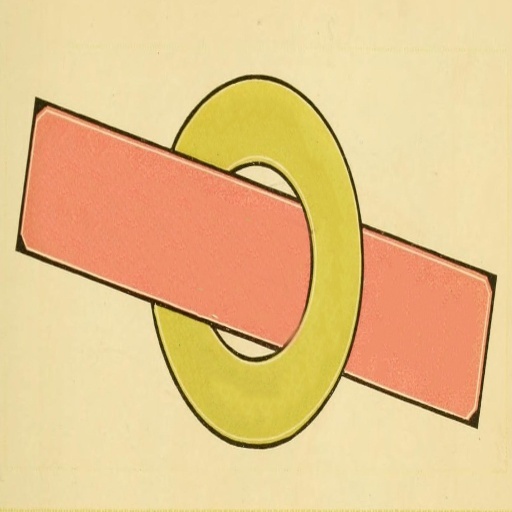}\vspace{4pt}
\end{minipage}}\vspace{2.5pt}
\hspace*{-8pt}
\subfigure[Ours]{
\begin{minipage}[t]{0.241\textwidth}
    \includegraphics[height=0.85\textwidth, width=1\textwidth]{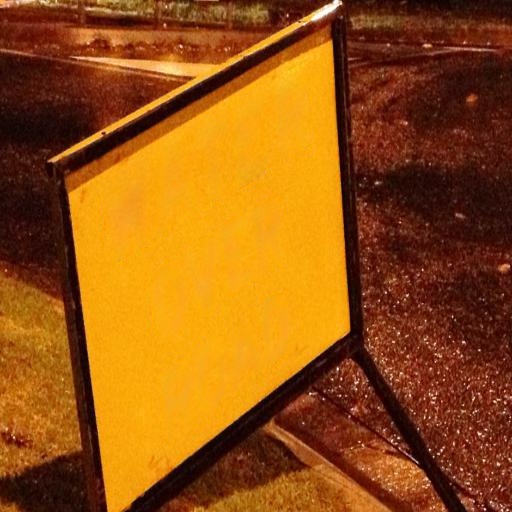}\vspace{2.5pt}
    \includegraphics[height=0.85\textwidth, width=1\textwidth]{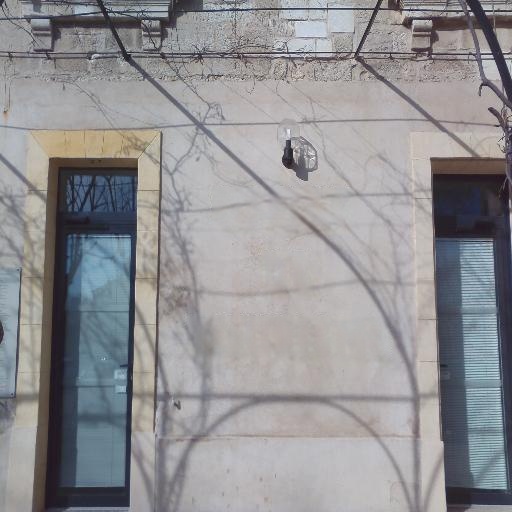}\vspace{2.5pt}
    \includegraphics[height=0.85\textwidth, width=1\textwidth]{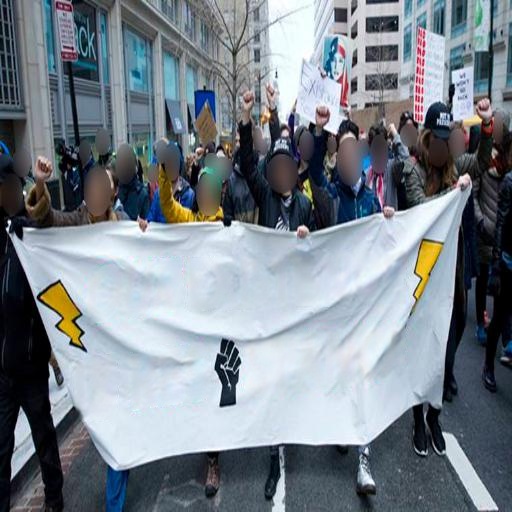}\vspace{2.5pt}
    \includegraphics[height=0.85\textwidth, width=1\textwidth]{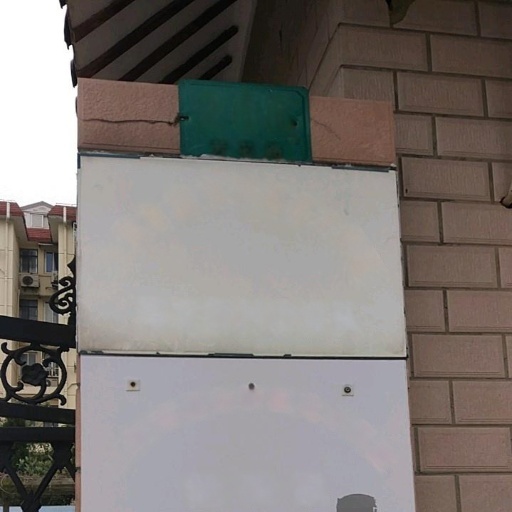}\vspace{2.5pt}
    \includegraphics[height=0.85\textwidth, width=1\textwidth]{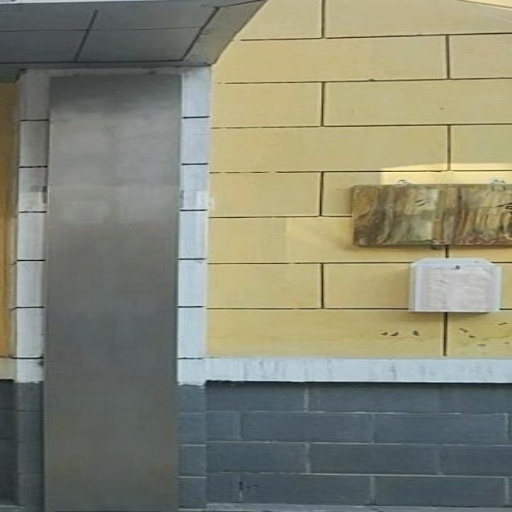}\vspace{2.5pt}
    \includegraphics[height=0.85\textwidth, width=1\textwidth]{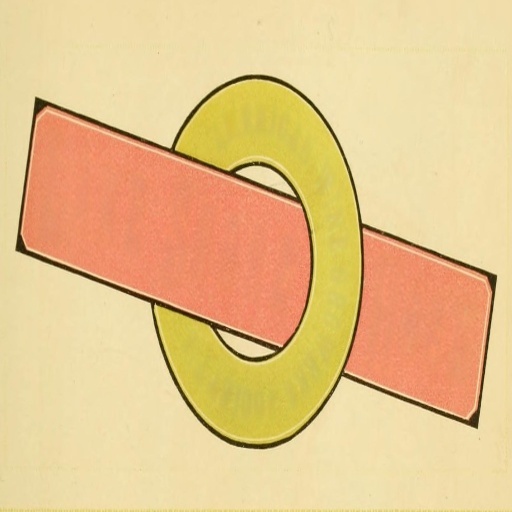}\vspace{4pt}
\end{minipage}}
\hspace*{-8pt}
\subfigure[Predicted text mask]{
\begin{minipage}[t]{0.241\textwidth}
    \setlength{\fboxrule}{0.4pt}
    \setlength{\fboxsep}{0pt}
    \fbox{\includegraphics[height=0.85\textwidth, width=1\textwidth]{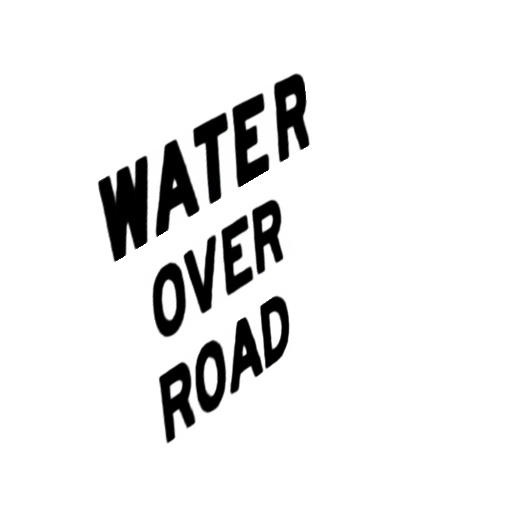}}\vspace{1.7pt}
    \fbox{\includegraphics[height=0.85\textwidth, width=1\textwidth]{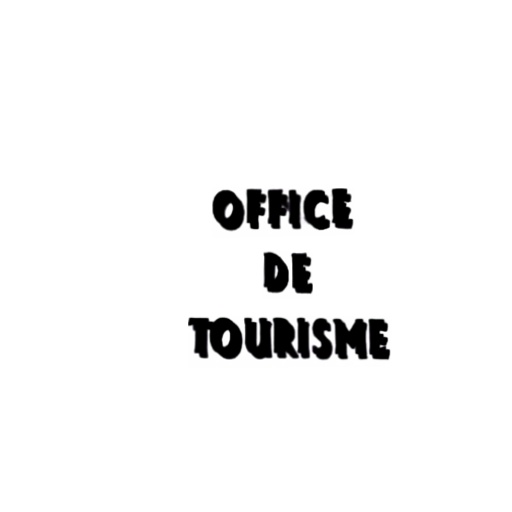}}\vspace{1.7pt}
    \fbox{\includegraphics[height=0.85\textwidth, width=1\textwidth]{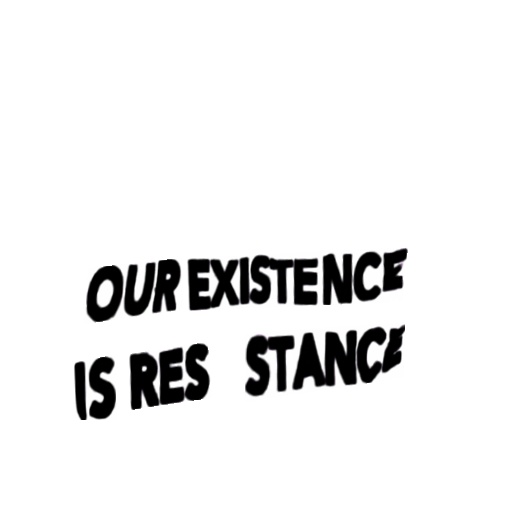}}\vspace{1.7pt}
    \fbox{\includegraphics[height=0.85\textwidth, width=1\textwidth]{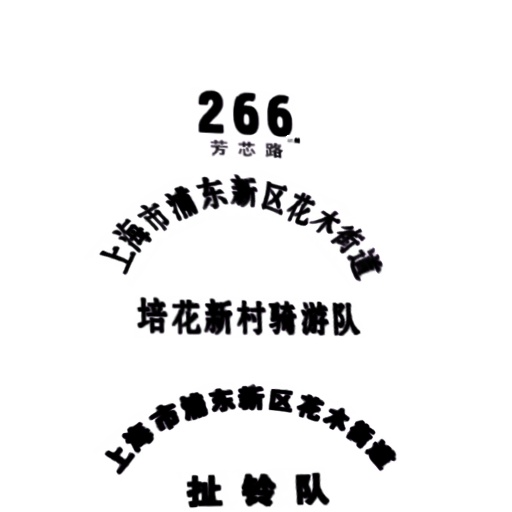}}\vspace{1.7pt}
    \fbox{\includegraphics[height=0.85\textwidth, width=1\textwidth]{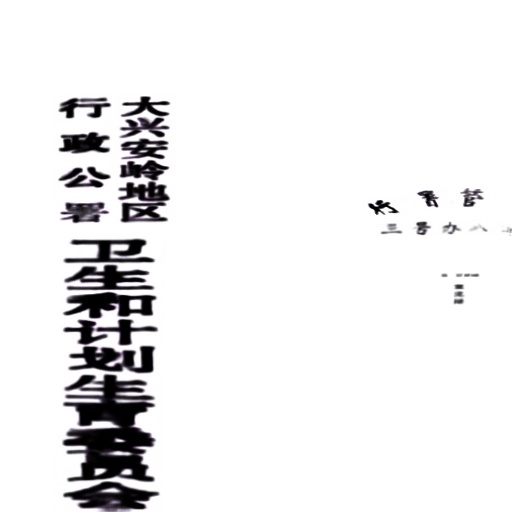}}\vspace{1.7pt}
    \fbox{\includegraphics[height=0.85\textwidth, width=1\textwidth]{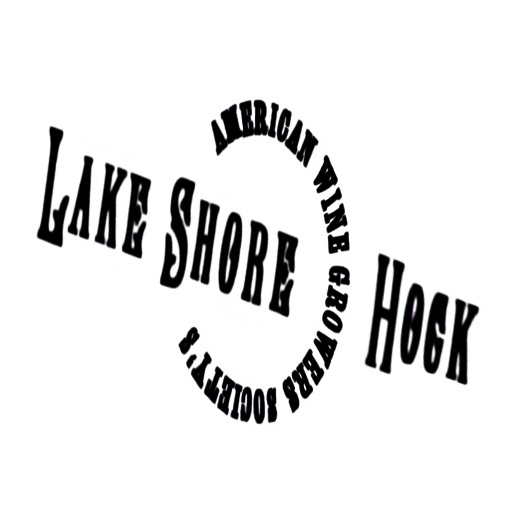}}\vspace{4pt}
\end{minipage}}
\vspace{-4pt}
\caption{Qualitative results of our method on SCUT-EnsText dataset. From left to right: input, ground truth, output of our method, and predicted text mask.} \label{tang8}
\end{figure*}

% % figure.9
% \begin{figure*}[!ht]
% \begin{minipage}{\linewidth}
% \includegraphics[width=\textwidth]{tang8.png}
% \leftline{\qquad\qquad\quad Input  \qquad\qquad\qquad\qquad\quad   Ground-truth \qquad\qquad\qquad\qquad\ Ours \qquad\qquad\qquad\qquad Predicted text mask}
% \end{minipage}
% \caption{Qualitative results of our method on SCUT-EnsText dataset. From left to right: input, ground truth, output of our method, and predicted text mask.} \label{tang8}
% \end{figure*}
% %

% table.2
\begin{table*}[!t]
\centering
\caption{Comparison Between Previous Scene Text-erasing Studies and Our Proposed Method on the SCUT-Syn and ICDAR2013 Datasets.}
\label{tab4.41}
\begin{tabular}{c|ccc|c|c|c|c}
\hline
\multirow{2}{*}{Method}                           & \multicolumn{3}{c|}{SCUT-Syn}                                                 & ICDAR2013  & \multirow{2}{*}{Parameters} & \multicolumn{1}{l|}{\multirow{2}{*}{Inference speed}} & \multirow{2}{*}{Input}            \\ \cline{2-5}
                                                  & \multicolumn{1}{c|}{PSNR↑} & \multicolumn{1}{c|}{SSIM(\%)↑} & MSE↓            & R↓         &                             & \multicolumn{1}{l|}{}                                 &                                   \\ \hline
Original images                                   & -                          & -                              & -               & 70.83      & -                           & -                                                     & -                                 \\
SceneTextEraser \cite{Nakamura2017}               & 14.68                      & 46.13                          & 0.7148          & 10.08      & -                           & -                                                     & Image                             \\
Pix2Pix \cite{Isolapixel2pixel2017}               & 25.60                      & 89.86                          & 0.2465          & 10.19      & 54.4M                       & \textbf{17ms}                                         & Image                             \\
EnsNet \cite{zhang_Ensnet_2019}                   & 37.36                      & 96.44                          & 0.0021          & 5.66       & 12.4M                       & 24ms                                                  & Image                             \\
MTRNet \cite{TursunMTRNet2019}                    & 29.71                      & 94.43                          & 0.0001          & 0.18       & 54.4M                       & -                                                     & Image(256$\times$256) + Text Mask \\
Weak Supervision \cite{Zdenekweaksupervision2020} & 37.44                      & 93.69                          & -               & 2.47       & 28.7+6.0M                   & 57+39ms                                               & Image(256$\times$256)             \\
MTRNet++ \cite{TursunMTRNetpp2020}                & 34.55                      & \textbf{98.45}                 & 0.0004          & -          & 18.7M                       & 37ms                                                  & Image(256$\times$256)             \\
EraseNet \cite{LiuEraseNet2020}                   & 38.32                      & 97.67                          & \textbf{0.0002} & -          & 19.7M                       & 34ms                                                  & Image                             \\ \hline
EAST \cite{zhou_east_2017} + Ours                 & 31.18                      & 95.93                          & 0.002           & 0.73       & 24.1+9.9M                   & 18+23$\sim$ms                                         & Image                             \\
Ours                                              & \textbf{38.60}             & 97.55                          & \textbf{0.0002} & \textbf{0} & \textbf{9.9M}               & 23$\sim$ms                                            & Image + BBox                      \\ \hline
\end{tabular}
\end{table*}

\subsection{Ablation Study} % 4.3
% \label{ssec:subhead}
In this section, we investigate the effectiveness of the different settings of the proposed model. The stroke mask prediction module (SMPM), skip connection (SC) between two modules, self-attention block (SA), partial convolutions (PConv), L1 loss, and dice loss are the focus of this study. The qualitative evaluation results on the SCUT-Syn and SCUT-EnsText datasets are presented in Table~\ref{tab4.3}, and some text-erasing samples are shown in Fig.~\ref{tang6}.
\begin{itemize}

\item[\textbullet ] \textbf{Stroke Mask Prediction Module} SMPM aims to provide the pixel-level information of text region as the hole for background inpainting Module (BIPM), so that the network can learn more from the valid features of the non-text region and suppress the text residue. The qualitative results are presented in Table~\ref{tab4.3}. Using the text mask can significantly improve the text erasing performance. It should be noted that the partial convolutional layers in the background inpainting module, function as normal convolutional layers without the mask image from the SMPM. 

\item[\textbullet ] \textbf{Skip Connection} Skip connection links and concatenates the low-resolution feature maps of the two modules to provide the features inside the text region for the decoder of the BIPM and to improve the accuracy and stability of text mask prediction. Table~\ref{tab4.3} implies that the skip connection between the two modules can improve the erasing quality of the image in both the SCUT-Syn and SCUT-EnsText datasets.

%table.4
\begin{table*}[!t]
\centering
\caption{Comparison Between State-of-the-art Inpainting Methods and Proposed Method on the SCUT-EnsText Dataset.}
\label{tab4.42}
\begin{tabular}{c|ccc|c|c|c|c}
\hline
\multirow{2}{*}{Method}          & \multicolumn{3}{c|}{SCUT-EnsText}                                             & \multirow{2}{*}{Parameters} & \multirow{2}{*}{Inference speed} & \multirow{2}{*}{Input}     & \multirow{2}{*}{Training dataset}                   \\ \cline{2-4}
                                 & \multicolumn{1}{c|}{PSNR↑} & \multicolumn{1}{c|}{SSIM(\%)↑} & MSE↓            &                             &                                  &                            &                                                     \\ \hline
LBAM \cite{xie_LBAM_2019}        & 36.21                      & 95.58                          & 0.0007          & 68.3M                       & \textbf{11ms}                    & Image(256×256) + Text Mask & Paris Street View \cite{DoerschParisstreetview2012} \\
RFR-Net \cite{li_recurrent_2020} & 36.95                      & 96.12                          & 0.0006          & 31.2M                       & 90ms                             & Image(256×256) + Text Mask & Paris Street View \cite{DoerschParisstreetview2012} \\
HiFill \cite{yi_HiFill_2020}     & 31.48                      & 94.17                          & 0.0021          & \textbf{2.7M}               & 28ms                             & Image + Text Mask          & Places2 \cite{zhou_places_2018}                     \\ \hline
Ours                             & \textbf{37.89}             & \textbf{97.02}                 & \textbf{0.0004} & 9.9M                        & 23$\sim$ms                      & Image(256×256) + BBox      & Improved Synth text                                 \\
Ours                             & \textbf{37.08}             & \textbf{96.54}                 & \textbf{0.0005} & 9.9M                        & 23$\sim$ms                       & Image + BBox               & Improved Synth text                                 \\ \hline
\end{tabular}
\end{table*}

\item[\textbullet ] \textbf{Self-Attention Block} GCblock adds channel-wise weights to the input feature maps taking into consideration the correspondences between feature maps and non-local features. To confirm the importance of the self-attention block, we trained our network without the SA block. In Table~\ref{tab4.3}, we see that the performance of our network decreases when the self-attention block is missing.

\item[\textbullet ] \textbf{Partial Convolutions} To evaluate the advantages of the partial convolutional layers, we also re-implemented our method without these layers. Table~\ref{tab4.3} lists the qualitative performance of the SCUT-Syn and SCUT-EnsText datasets. We observed that, compared to a network with partial convolutional layers, the network with partial convolutional layers performs better on SCUT-EnsText but worse on SCUT-Syn datasets. We believe that the discrepancies between synthetic data and real-world data are the cause for this difference in performance. As mentioned before, the reason for improving the Synth-text engine is that Poisson image editing retains some texture information of the background image when it blends the foreground text instances into a background image. However, most scene text instances in real-world images are not transparent. For the erasure result on SCUT-Syn datasets, compared with directly extracting features using normal convolution, using partial convolution and $F_{m}$ feature concatenation to split the features inside and outside the text region is a relatively inefficient approach for transparent text erasure. However, when facing the real-world data such as SCUT-EnsText, using partial convolution can achieve better performance than normal convolution. 

\item[\textbullet ] \textbf{L1 Loss and Dice Loss} To confirm the contribution of these two losses in SMPM, we trained our whole network with single L1 loss or dice loss and compared the final erasure result with all. In Table~\ref{tab4.3}, we observe that our network with the L1 loss and dice loss combination achieved better results on the real-world dataset. 

\end{itemize}

% table.5
\begin{table*}[!t]
\centering
\caption{Comparison Between State-of-the-art Scene Text-erasing Methods and Proposed Method on the SCUT-EnsText Dataset. Re: We Re-implemented this Method.}
\label{tab4.43}
\begin{tabular}{c|ccc|c|c|c|c}
\hline
\multirow{2}{*}{Method}                        & \multicolumn{3}{c|}{Qualitative eval}                                         & Quantitative eval & \multirow{2}{*}{Parameters} & \multirow{2}{*}{Inference speed} & \multirow{2}{*}{Input} \\ \cline{2-5}
                                               & \multicolumn{1}{c|}{PSNR↑} & \multicolumn{1}{c|}{SSIM(\%)↑} & MSE↓            & R↓                &                             &                                  &                        \\ \hline
Original   images                              & -                          & -                              & -               & 69.5              & -                           & -                                & -                      \\
SceneTextEraser \cite{Nakamura2017}            & 25.47                      & 90.14                          & 0.0047          & 5.9               & -                           & -                                & Image                  \\
EnsNet \cite{zhang_Ensnet_2019}                & 29.54                      & 92.74                          & 0.0024          & 32.8              & \textbf{12.4M}              & \textbf{24ms}                    & Image                  \\
EraseNet \cite{LiuEraseNet2020}                & 32.30                      & 95.42                          & 0.0015          & 4.6               & 19.7M                       & 34ms                             & Image                  \\
EraseNet \cite{LiuEraseNet2020} (Re)           & 32.64                      & 95.13                          & 0.0017          & 5.4               & 19.7M                       & 34ms                             & Image                  \\
DB-ResNet-50 \cite{LiaoDB2020} + EraseNet (Re) & 32.70                      & 95.18                          & 0.0021          & 12.3              & 26.1+19.7M                  & 37+34ms                          & Image                  \\
CRAFT \cite{BaekCRAFT2019} + EraseNet (Re)     & 34.15                      & 95.66                          & 0.0015          & 7.3               & 20.7+19.7M                  & 120+34ms                         & Image                  \\ \hline
DB-ResNet-18 \cite{LiaoDB2020} + Ours          & 33.17                      & 95.44                          & 0.0020          & 10.3              & 12.6+9.9M                   & 17+27$\sim$ms                    & Image                  \\
DB-ResNet-50 \cite{LiaoDB2020} + Ours          & 33.54                      & 95.57                          & 0.0018          & 10.5              & 26.1+9.9M                   & 37+27$\sim$ms                    & Image                  \\
CRAFT \cite{BaekCRAFT2019} + Ours              & \textbf{35.34}             & \textbf{96.24}                 & \textbf{0.0009} & \textbf{3.6}      & 20.7+9.9M                   & 120+27$\sim$ms                   & Image                  \\ \hline
\end{tabular}
\end{table*}

\subsection{Comparison With State-of-the-Art Methods} % 4.4
To evaluate the performance of our proposed method, we compared it with recent state-of-the-art methods on the SCUT-Syn, ICDAR2013, and SCUT-EnsText datasets. For the SCUT-Syn and ICDAR2013 datasets, the results of SceneTextEraser, Pix2Pix, and EnsNet were implemented and reported by Zhang \textit{et al.} \cite{zhang_Ensnet_2019}. The results of MTRNet \cite{TursunMTRNet2019}, weak supervision \cite{Zdenekweaksupervision2020}, MTRNet++ \cite{TursunMTRNetpp2020}, and EraseNet \cite{LiuEraseNet2020} were collected from official reports. If there is no specific description, the resolution of the input image is $512\times512$. Table~\ref{tab4.41} displays the results for the SCUT-Syn \cite{zhang_Ensnet_2019} and ICDAR2013 \cite{KaratzasICDAR2013} datasets. Our proposed method achieves the highest PSNR on the SCUT-Syn dataset and the lowest recall on the ICDAR2013 dataset when the bounding boxes are provided. Some text-erasing examples on the SCUT-Syn dataset are shown in Fig.~\ref{tang7}. Our method achieves lower recall on the ICDAR2013 dataset when the detection result of EAST \cite{zhou_east_2017} is used to provide text location. Here, R represents recall, which is the detection result of the EAST under the ICDAR2013 evaluation protocol. We believe the reason why MTRNet++, EraseNet, and EnsNet could generate higher SSIM images on the SCUT-Syn dataset is because most test images are included in the training set, and share the same background images with training images when they are generated. The EAST and our model were not trained on the SCUT-Syn training set, thereby resulting in a lower PSNR and SSIM. We believe that this dataset cannot fully reflect the generalization ability of a network when it is used for training. Our text-erasing network is lightweight with only 9.9 million trainable parameters. For a fair comparison of the inference speed, we tested all methods using a single 1080ti GPU and an AMD Ryzen7 3700X @ 3.6GHz CPU with the original input size of the networks. The inference time of our method consisted of the time cost of the network forward, pre-processing, and post-processing. The time cost of pre-processing and post-processing was approximately 4 ms in the case of using perspective transformation and 76 ms in the case of using off-the-shell thin-plate-spline function in OpenCV.

% % figure.10
\begin{figure*}[!ht]
\centering
\subfigure[Input]{
\begin{minipage}[t]{0.241\textwidth}
    \includegraphics[height=0.85\textwidth, width=1\textwidth]{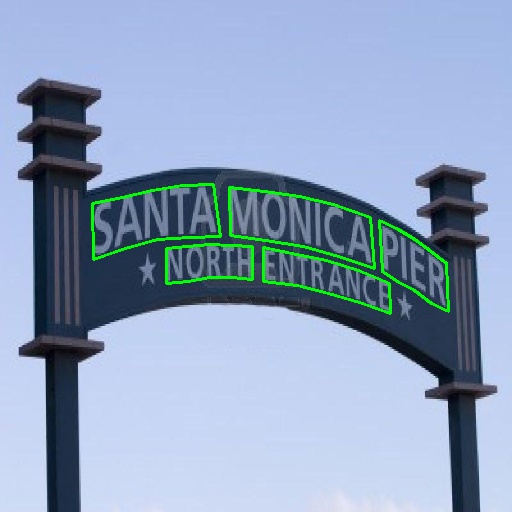}\vspace{2.5pt}
    \includegraphics[height=0.85\textwidth, width=1\textwidth]{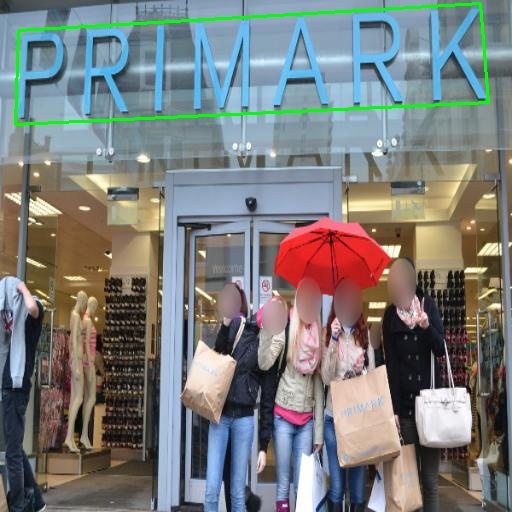}\vspace{2.5pt}
    \includegraphics[height=0.85\textwidth, width=1\textwidth]{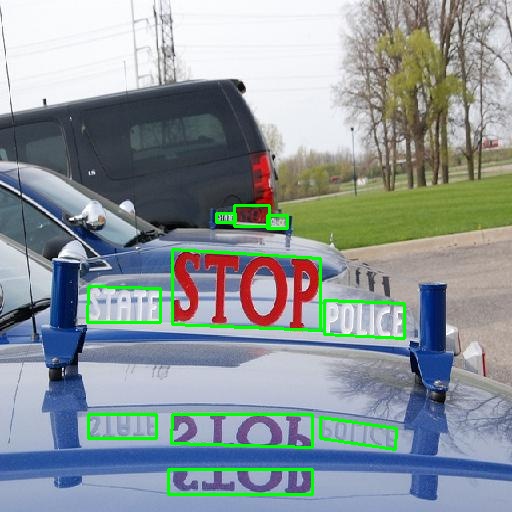}\vspace{2.5pt}
    \includegraphics[height=0.85\textwidth, width=1\textwidth]{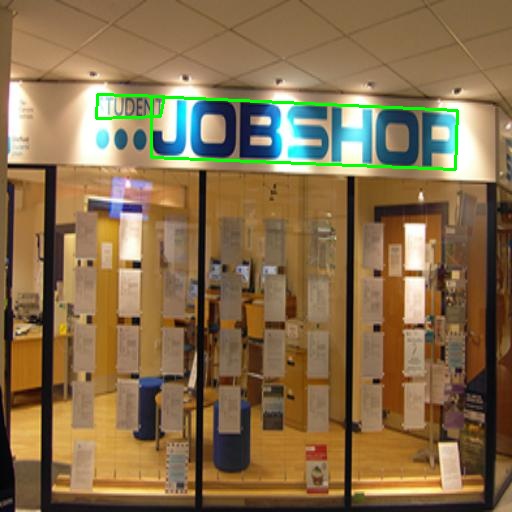}\vspace{4pt}
\end{minipage}}
\hspace*{-8pt}
\subfigure[Ground-truth]{
\begin{minipage}[t]{0.241\textwidth}
    \includegraphics[height=0.85\textwidth, width=1\textwidth]{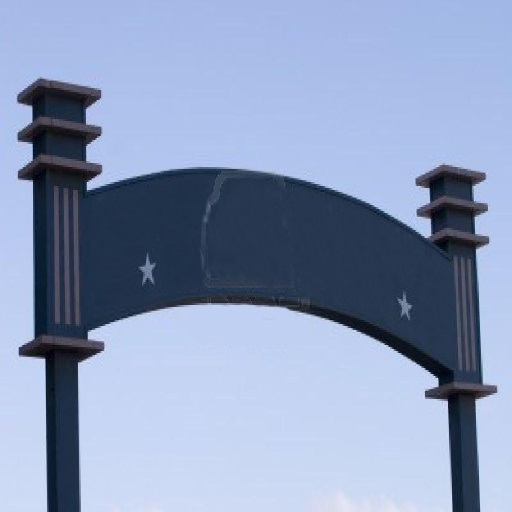}\vspace{2.5pt}
    \includegraphics[height=0.85\textwidth, width=1\textwidth]{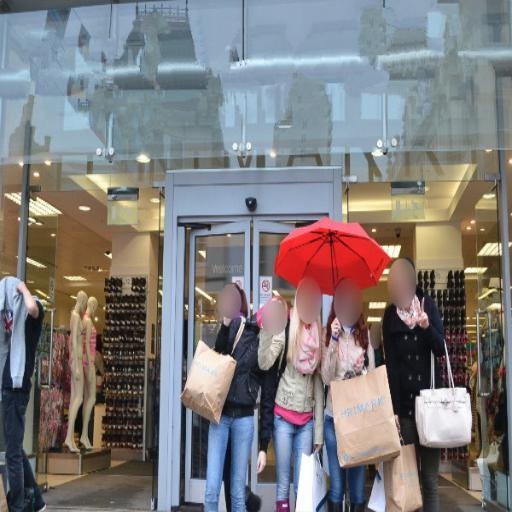}\vspace{2.5pt}
    \includegraphics[height=0.85\textwidth, width=1\textwidth]{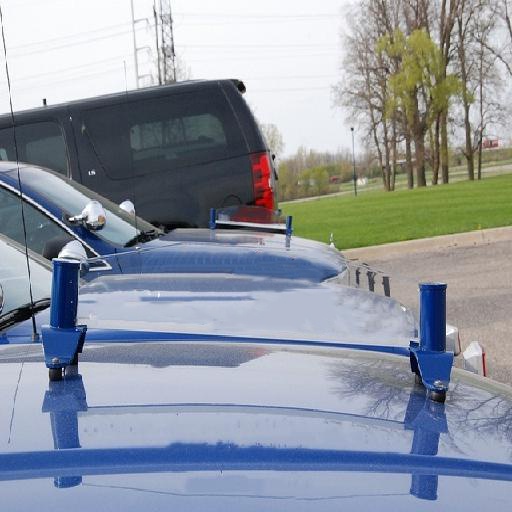}\vspace{2.5pt}
    \includegraphics[height=0.85\textwidth, width=1\textwidth]{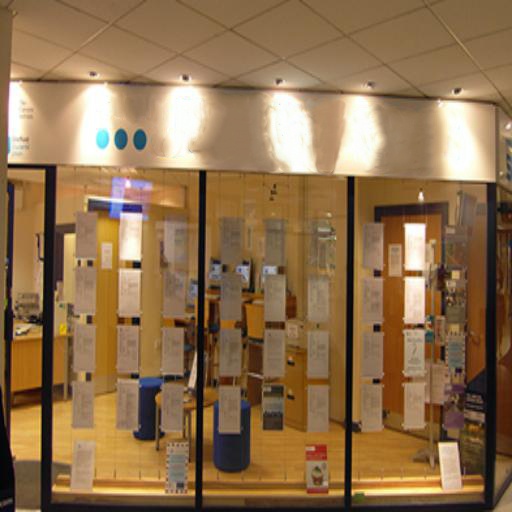}\vspace{4pt}
\end{minipage}}\vspace{2.5pt}
\hspace*{-8pt}
\subfigure[Ours]{
\begin{minipage}[t]{0.241\textwidth}
    \includegraphics[height=0.85\textwidth, width=1\textwidth]{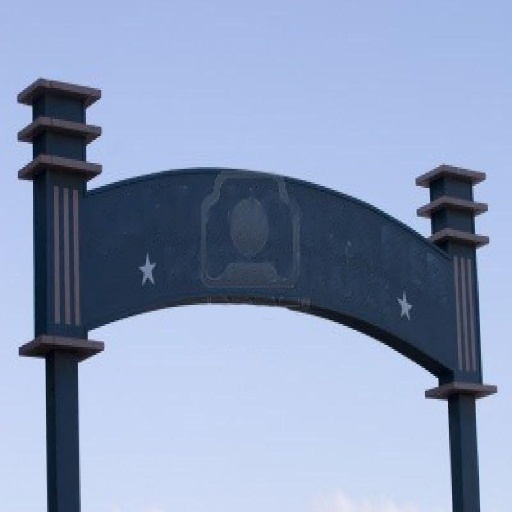}\vspace{2.5pt}
    \includegraphics[height=0.85\textwidth, width=1\textwidth]{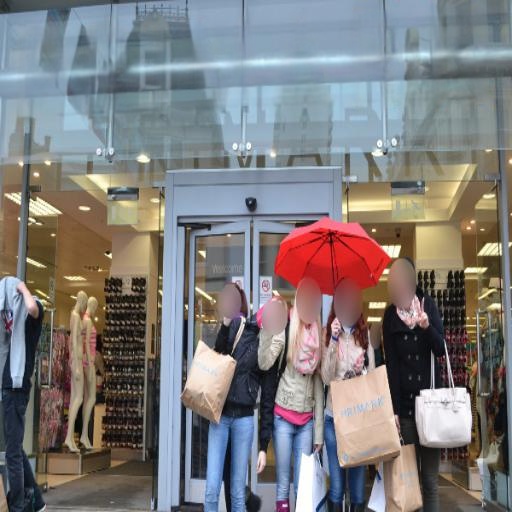}\vspace{2.5pt}
    \includegraphics[height=0.85\textwidth, width=1\textwidth]{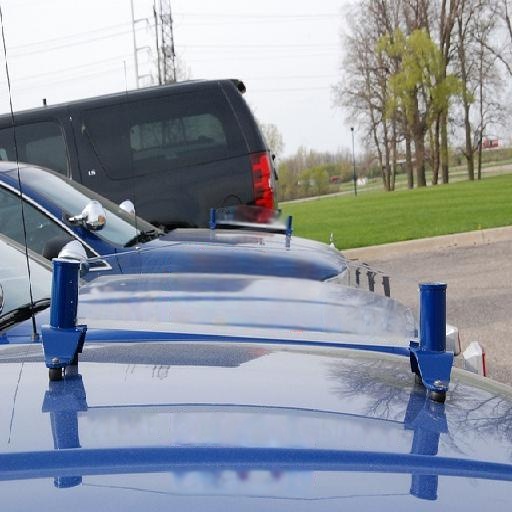}\vspace{2.5pt}
    \includegraphics[height=0.85\textwidth, width=1\textwidth]{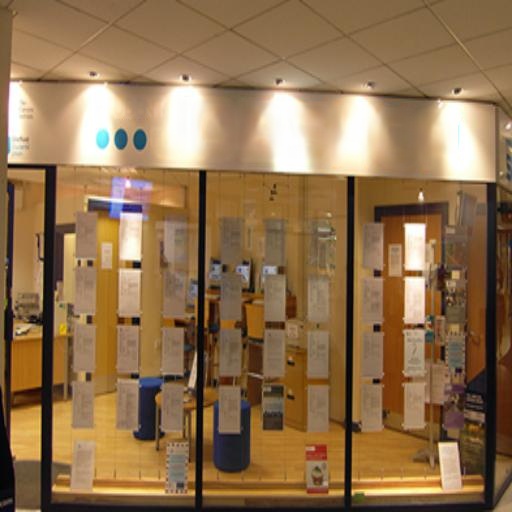}\vspace{4pt}
\end{minipage}}
\hspace*{-8pt}
\subfigure[Predicted text mask]{
\begin{minipage}[t]{0.241\textwidth}
    \setlength{\fboxrule}{0.4pt}
    \setlength{\fboxsep}{0pt}
    \fbox{\includegraphics[height=0.85\textwidth, width=1\textwidth]{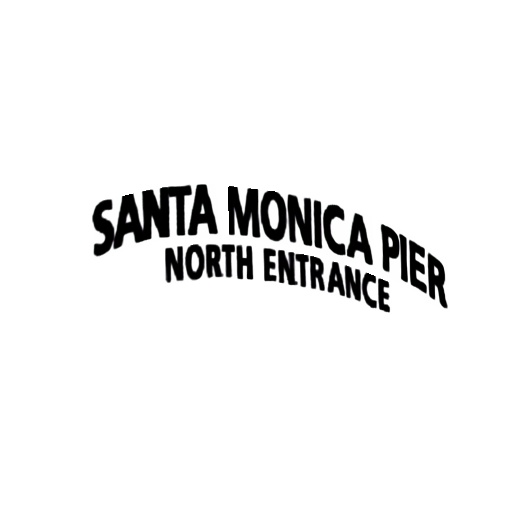}}\vspace{1.7pt}
    \fbox{\includegraphics[height=0.85\textwidth, width=1\textwidth]{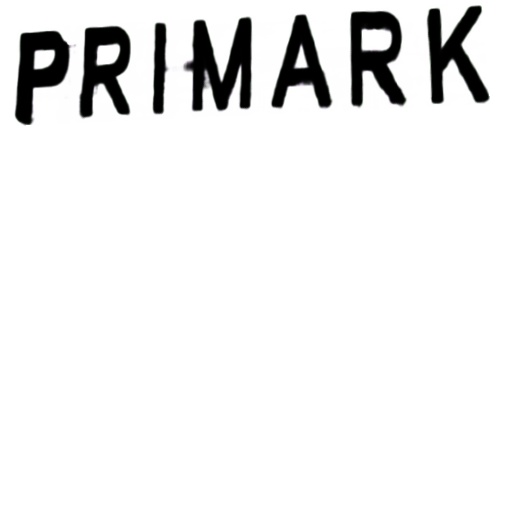}}\vspace{1.7pt}
    \fbox{\includegraphics[height=0.85\textwidth, width=1\textwidth]{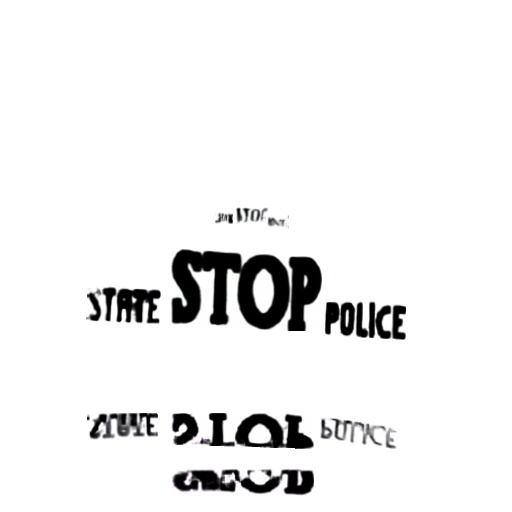}}\vspace{1.7pt}
    \fbox{\includegraphics[height=0.85\textwidth, width=1\textwidth]{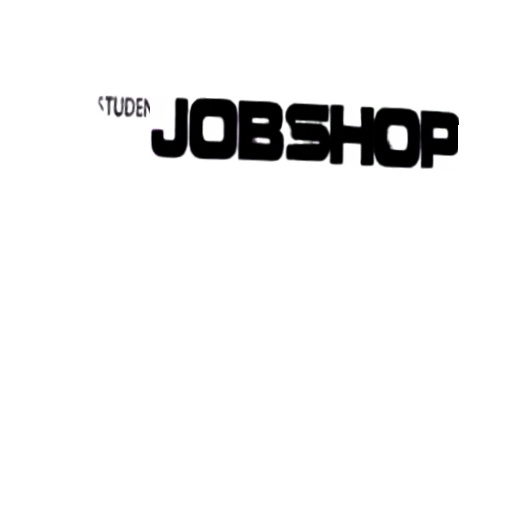}}\vspace{4pt}
\end{minipage}}
\vspace{-4pt}
\caption{Our method can retain more detailed background information and restore the background texture. From left to right: input image and text bounding boxes, ground truth, output of our method, and predicted text mask.} \label{tang10}
\end{figure*}

% % figure.10
% \begin{figure*}[!t]
% \begin{minipage}{\linewidth}
% \includegraphics[width=\textwidth]{tang10.png}
% \leftline{\qquad\qquad\quad Input  \qquad\qquad\qquad\qquad\quad   Ground-truth \qquad\qquad\qquad\qquad\ Ours \qquad\qquad\qquad\qquad Predicted text mask}
% \end{minipage}
% \caption{Our method is able to retain more detailed background information and restore the background texture. From left to right: input image and text bounding boxes, ground truth, our output, and predicted text mask.} \label{tang10}
% \end{figure*}
% %

% fig11
\begin{figure*}[!ht]
\centering
\subfigure[Input]{
\begin{minipage}[t]{0.241\textwidth}
    \includegraphics[height=0.85\textwidth, width=1\textwidth]{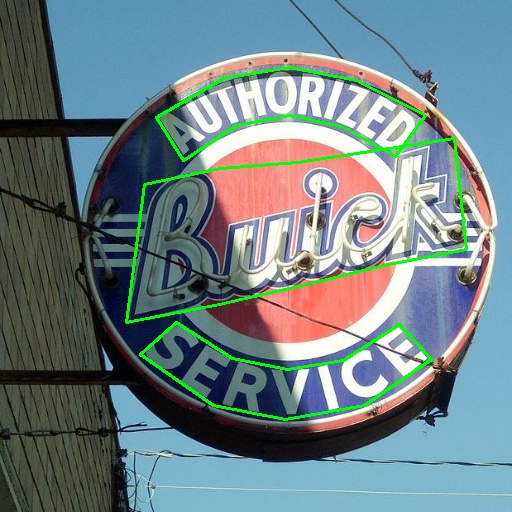}\vspace{2.5pt}
    \includegraphics[height=0.85\textwidth, width=1\textwidth]{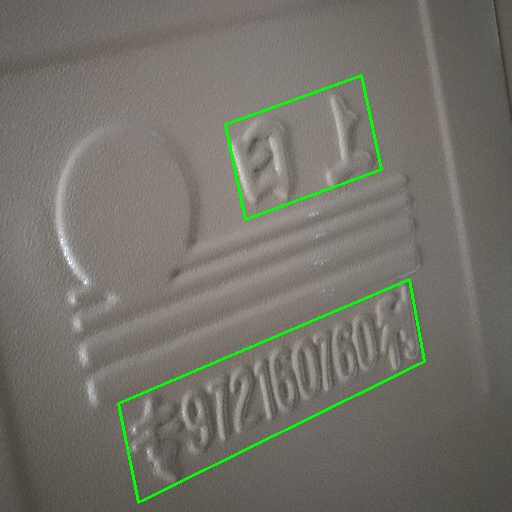}\vspace{2.5pt}
    \includegraphics[height=0.85\textwidth, width=1\textwidth]{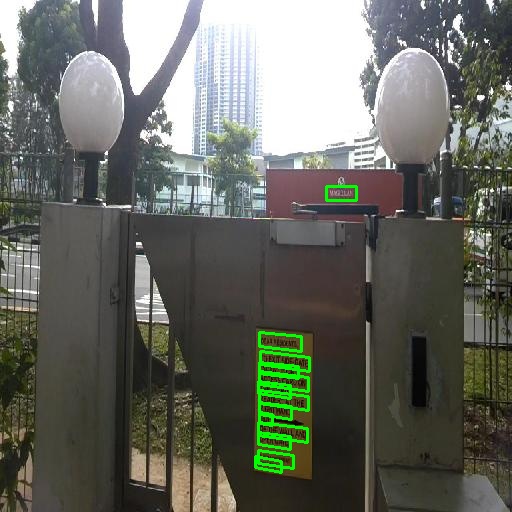}\vspace{4pt}
\end{minipage}}
\hspace*{-8pt}
\subfigure[Ground-truth]{
\begin{minipage}[t]{0.241\textwidth}
    \includegraphics[height=0.85\textwidth, width=1\textwidth]{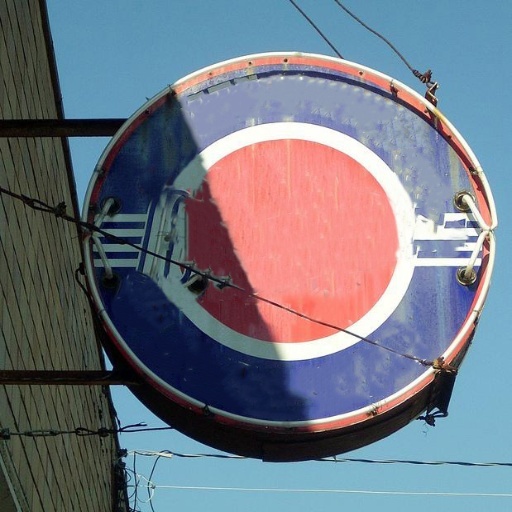}\vspace{2.5pt}
    \includegraphics[height=0.85\textwidth, width=1\textwidth]{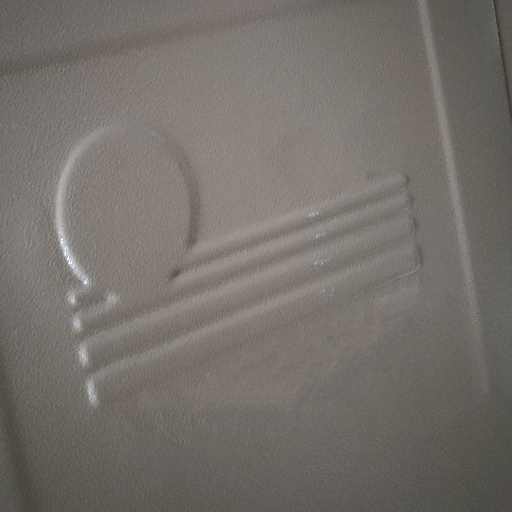}\vspace{2.5pt}
    \includegraphics[height=0.85\textwidth, width=1\textwidth]{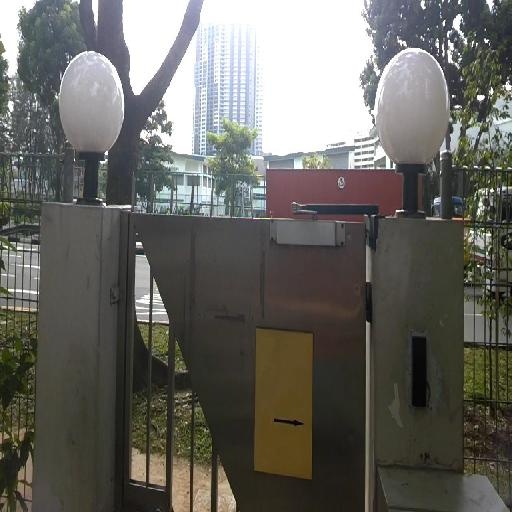}\vspace{4pt}
\end{minipage}}\vspace{2.5pt}
\hspace*{-8pt}
\subfigure[Ours]{
\begin{minipage}[t]{0.241\textwidth}
    \includegraphics[height=0.85\textwidth, width=1\textwidth]{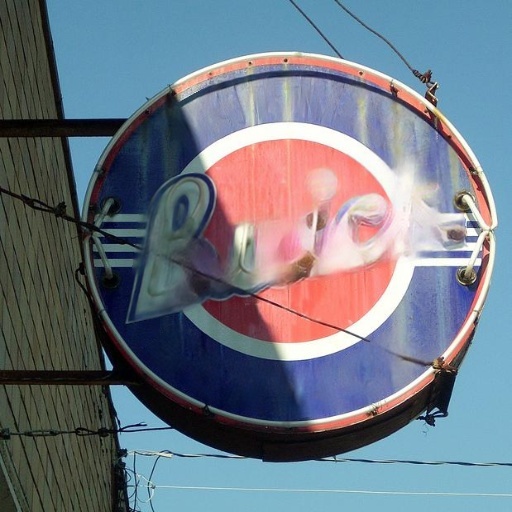}\vspace{2.5pt}
    \includegraphics[height=0.85\textwidth, width=1\textwidth]{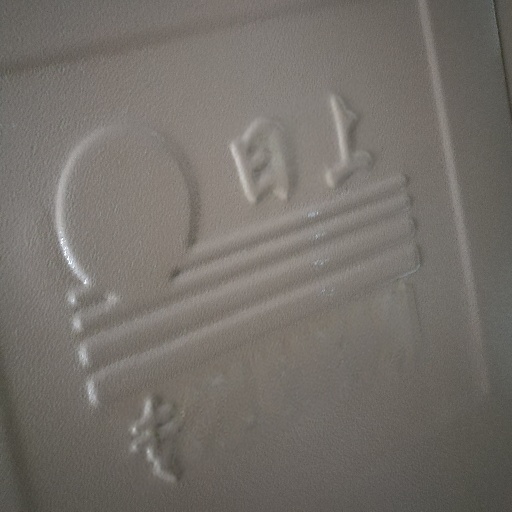}\vspace{2.5pt}
    \includegraphics[height=0.85\textwidth, width=1\textwidth]{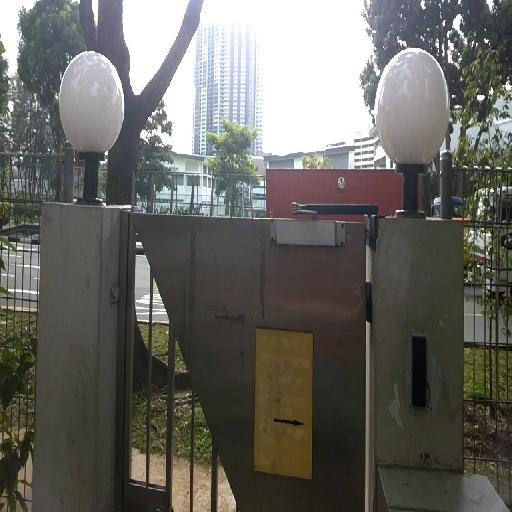}\vspace{4pt}
\end{minipage}}
\hspace*{-8pt}
\subfigure[Predicted text mask]{
\begin{minipage}[t]{0.241\textwidth}
    \setlength{\fboxrule}{0.4pt}
    \setlength{\fboxsep}{0pt}
    \fbox{\includegraphics[height=0.85\textwidth, width=1\textwidth]{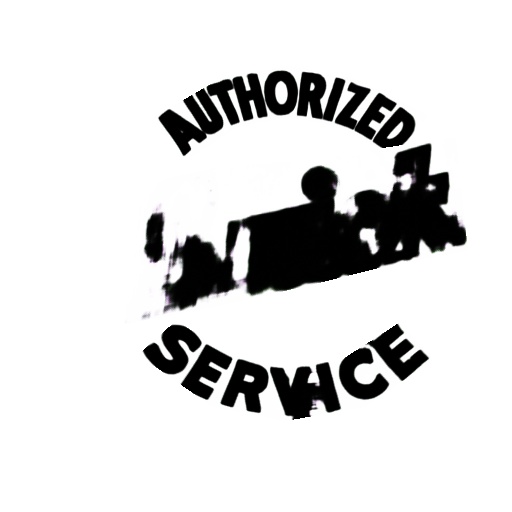}}\vspace{1.7pt}
    \fbox{\includegraphics[height=0.85\textwidth, width=1\textwidth]{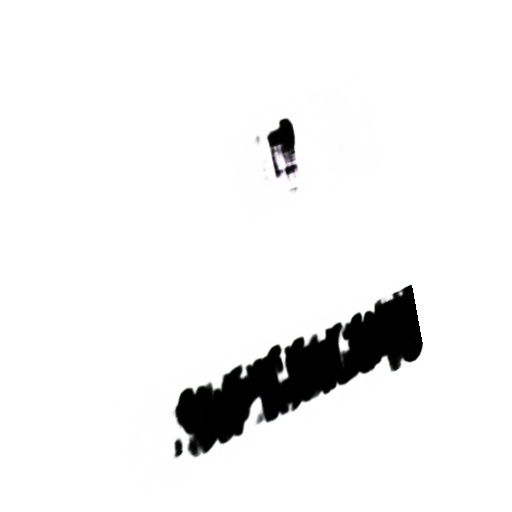}}\vspace{1.7pt}
    \fbox{\includegraphics[height=0.85\textwidth, width=1\textwidth]{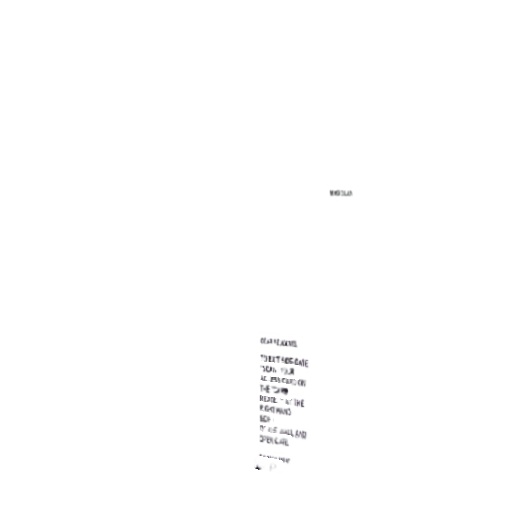}}\vspace{4pt}
\end{minipage}}
\vspace{-4pt}
\caption{Some failure cases of our method. From left to right: input image and text bounding boxes, ground truth, output of our method, and predicted text mask.} \label{tang11}
\end{figure*}

For the SCUT-EnsText dataset, we compared our method with state-of-the-art image inpainting methods and scene text erasing methods. The results of the comparison with the inpainting methods of the SCUT-EnsText dataset are shown in Table~\ref{tab4.42}. Our scene text erasing method achieves excellent results in image quality when text bounding box information is provided.
We made some revisions to the original text location annotation because we observed some unmatched cases between the location of the erased text and the text bounding box of the ground truth. We selected three state-of-the-art image inpainting methods: LBAM \cite{xie_LBAM_2019}, RFR-Net \cite{li_recurrent_2020}, and HiFill \cite{yi_HiFill_2020}. The LBAM \cite{xie_LBAM_2019} and RFR-Net \cite{li_recurrent_2020} models were pretrained on the Paris Street View dataset \cite{DoerschParisstreetview2012}, and the HiFill \cite{yi_HiFill_2020} model was pretrained on the Places2 dataset \cite{zhou_places_2018}. For a fair comparison with the pretrained inpainting methods, we generated the hole mask directly from our revised text location annotations and resized the input images to the same size because some pretrained inpainting models only work at a resolution of 256$\times$256. Our method generated images with higher quality than that of the state-of-the-art image inpainting methods under the same conditions. Some samples are presented for visual quality comparison in Fig.~\ref{tang9}. We observed that there are certain unusual textures or artifacts in the text-erased images inpainted via pretrained image inpainting methods, causing unnatural erasure results. We also found the inpainting logic of HiFill \cite{yi_HiFill_2020}, a preference for restoring the hole region with further background information is displayed, leading to worse results than those obtained using the other two methods. Although this model is pretrained on the Places2 \cite{zhou_places_2018} dataset, which contains many indoor and urban views, it still presents a serious domain shift problem when facing scene-text-erasing tasks.

In addition, we used our method with a pretrained scene text detector as a two-step automatic scene text eraser. In our experiment, we used DB \cite{LiaoDB2020} and CRAFT \cite{BaekCRAFT2019} for scene text detection and produced arbitrary quadrilateral bounding boxes as the input for our method. DB-ResNet-18 and DB-ResNet-50 were pretrained in the SynthText and ICDAR 2015 datasets with the box threshold set at 0.3. CRAFT was pretrained on the SynthText, ICDAR 2013, and MLT 2017 datasets, and the text threshold was set to 0.6. We compared our proposed methods with previous scene text erasing methods on the SCUT-EnsText dataset. For a fair comparison with one-step methods, such as EraseNet, we carefully re-implemented EraseNet and replaced the background regions of the output of EraseNet with the original images according to the detection results of the scene text detector. The results are shown in Table~\ref{tab4.43} and imply that the results of the first detection followed by inpainting via our method are significantly superior to those of existing state-of-the-art one-step methods in both qualitative and quantitative evaluations. Here, R denotes the recall, which is the detection result of the CRAFT \cite{BaekCRAFT2019} under the ICDAR2015 evaluation \cite{KaratzasICDAR2015} protocol. Overall, compared with a single EraseNet, using an auxiliary scene text detector with the one-step method could help increase the quality of the output images by replacing excessive erasure of text-free areas with content from original images. On the other hand, this operation could also restore some text-erased instances to their original status because of the limited detection ability of the text detector, resulting in an increase in recall. Some qualitative results of our method on the SCUT-EnsText dataset are shown in Fig.~\ref{tang8}. Our method can clearly remove the text region, regardless of whether the text instances have varying shapes, fonts, or illumination conditions. Nonetheless, compared with one-step end-to-end scene text erasing methods, the inference speed of our two-step cropped-images-based pipeline is relatively slow, as shown in Table~\ref{tab4.43}.

% %
% \begin{figure*}[!t]
% \begin{minipage}{\linewidth}
% \includegraphics[width=\textwidth]{tang11.png}
% \leftline{\qquad\qquad\quad Input  \qquad\qquad\qquad\qquad\quad   Ground-truth \qquad\qquad\qquad\qquad\ Ours \qquad\qquad\qquad\qquad Predicted text mask}
% \end{minipage}
% \caption{Some failure cases of our method. From left to right: input image and text bounding boxes, ground truth, output, and predicted text mask.} \label{tang11}
% \end{figure*}
% %

\subsection{Discussion} % 4.5
From the experimental results, we observe that the model trained using our improved synthetic text dataset displays a different inpainting logic than that of the annotations of the real-world scene-text removal dataset, which was manually edited using Photoshop. As shown in Fig.~\ref{tang10}, our method can erase text while retaining more detailed background information of the image. In addition, owing to the model design and a large amount of training data, our method was able to reconstruct some background texture for better visual perception.
However, because of the limited text style of the synthetic text engine, the method failed during text erasing in case of text oriented in special shape or the stereoscopic text under complicated illumination conditions, as depicted in Fig.~\ref{tang11}. Because we propose erasing word-level text in cropped images, our method encountered some difficulty in erasing small text instances in the images. As there are always JPEG boundary artifacts surrounding the text edge, our model cannot significantly discriminate whether the pixel near the text edge belongs to the background or to the artifact. Borrowing features from artifact regions can result in a text region being inpainted by strange colors, yielding poor inpainting results. 

When using scene text detector working with our method, the quality of the generated text bounding boxes also affects the quality of text erasure. Text instances partially covered by bounding boxes can result in residues of text outside boxes, because in our experimental setting, only the region inside the text bounding boxes are replaced by the output of our network. Moreover, curved text, bounded by poor quality polygons, will be heavily distorted after TPS transformation, which makes the prediction of stroke mask more difficult, resulting in poor text erasure. In contrast, curved text, with good polygon annotation, usually leads to high-quality erasing. In our evaluation, we found that using perspective transformation and quadrilateral bounding boxes is the safest way to maintain high-quality erasure on average.

\section{Conclusion} %5
\label{Conclusion}
In this study, we propose a novel scene text erasing method that addresses the weak text location problem of one-step methods and the domain shift problem of using inpainting models pretrained on street view or Places datasets. To this end, our model was trained using only our improved synthetic text dataset. The model inpaints the text region based on a predicted text stroke mask derived from cropped text images, whereby more background information can be preserved. By utilizing a stroke mask prediction module, partial convolution layers, an attention block in the background inpainting module, and skip connection between two modules, our method can reasonably erase scene text with texture restoration. Using a pretrained scene text detector to provide text location information, our model can function as an automatic scene text eraser to remove text from the wild. Overall, our experimental results show that our method performs better than existing state-of-the-art methods.

% or


% \appendices

% % you can choose not to have a title for an appendix
% % if you want by leaving the argument blank
% \section{}

% % use section* for acknowledgment
% \section*{Acknowledgment}


\ifCLASSOPTIONcaptionsoff
  \newpage
\fi



\bibliographystyle{IEEEtran}
% argument is your BibTeX string definitions and bibliography database(, s)
\bibliography{new_ref}

\begin{thebibliography}{10}
\providecommand{\url}[1]{#1}
\csname url@samestyle\endcsname
\providecommand{\newblock}{\relax}
\providecommand{\bibinfo}[2]{#2}
\providecommand{\BIBentrySTDinterwordspacing}{\spaceskip=0pt\relax}
\providecommand{\BIBentryALTinterwordstretchfactor}{4}
\providecommand{\BIBentryALTinterwordspacing}{\spaceskip=\fontdimen2\font plus
\BIBentryALTinterwordstretchfactor\fontdimen3\font minus
  \fontdimen4\font\relax}
\providecommand{\BIBforeignlanguage}[2]{{%
\expandafter\ifx\csname l@#1\endcsname\relax
\typeout{** WARNING: IEEEtran.bst: No hyphenation pattern has been}%
\typeout{** loaded for the language `#1'. Using the pattern for}%
\typeout{** the default language instead.}%
\else
\language=\csname l@#1\endcsname
\fi
#2}}
\providecommand{\BIBdecl}{\relax}
\BIBdecl

\bibitem{Nakamura2017}
T.~Nakamura, A.~Zhu, K.~Yanai, and S.~Uchida, ``{Scene Text Eraser},'' in
  \emph{Proc. Int. Conf. Doc. Anal. Recognit.}, vol.~1, 2017, pp. 832--837.

\bibitem{TursunMTRNetpp2020}
O.~Tursun, S.~Denman, R.~Zeng, S.~Sivapalan, S.~Sridharan, and C.~Fookes,
  ``{MTRNet++: One-stage mask-based scene text eraser},'' \emph{Comput. Vis.
  Image Underst.}, vol. 201, p. 103066, 2020.

\bibitem{zhang_Ensnet_2019}
S.~Zhang, Y.~Liu, L.~Jin, Y.~Huang, and S.~Lai, ``{EnsNet: Ensconce Text in the
  Wild},'' in \emph{Proc. AAAI Conf. Artif. Intell.}, vol.~33, 2019, pp.
  801--808.

\bibitem{LiuEraseNet2020}
C.~Liu, Y.~Liu, L.~Jin, S.~Zhang, C.~Luo, and Y.~Wang, ``{EraseNet: End-to-End
  Text Removal in the Wild},'' \emph{IEEE Trans. Image Process.}, vol.~29, pp.
  8760--8775, 2020.

\bibitem{gupta_synthetic_2016}
A.~Gupta, A.~Vedaldi, and A.~Zisserman, ``{Synthetic Data for Text Localisation
  in Natural Images},'' in \emph{Proc. IEEE Conf. Comput. Vis. Pattern
  Recognit.}, 2016, pp. 2315--2324.

\bibitem{Biancascaded2020}
X.~Bian, C.~Wang, W.~Quan, J.~Ye, X.~Zhang, and D.-M. Yan, ``{Scene text
  removal via cascaded text stroke detection and erasing},'' \emph{arXiv},
  2020.

\bibitem{TursunMTRNet2019}
O.~Tursun, R.~Zeng, S.~Denman, S.~Sivapalan, S.~Sridharan, and C.~Fookes,
  ``{MTRNet: A Generic Scene Text Eraser},'' in \emph{Proc. Int. Conf. Doc.
  Anal. Recognit.}, 2019, pp. 39--44.

\bibitem{Zdenekweaksupervision2020}
J.~Zdenek and H.~Nakayama, ``{Erasing Scene Text with Weak Supervision},'' in
  \emph{Proc. IEEE Winter Conf. Appl. Comput. Vis.}, 2020, pp. 2227--2235.

\bibitem{DoerschParisstreetview2012}
C.~Doersch, S.~Singh, A.~Gupta, J.~Sivic, and A.~A. Efros, ``{What makes Paris
  look like Paris?}'' \emph{ACM Trans. Graph.}, vol.~31, no.~4, pp. 1--9, 2012.

\bibitem{imageNet2015}
O.~Russakovsky, J.~Deng, H.~Su, J.~Krause, S.~Satheesh, S.~Ma, Z.~Huang,
  A.~Karpathy, A.~Khosla, M.~Bernstein, A.~C. Berg, and L.~Fei-Fei, ``{ImageNet
  Large Scale Visual Recognition Challenge},'' \emph{Int. J. Comput. Vis.},
  vol. 115, no.~3, pp. 211--252, 2015.

\bibitem{KaratzasICDAR2013}
D.~Karatzas, F.~Shafait, S.~Uchida, M.~Iwamura, L.~G.~I. Bigorda, S.~R. Mestre,
  J.~Mas, D.~F. Mota, J.~A. Almazan, and L.~P. de~las Heras, ``{ICDAR 2013
  Robust Reading Competition},'' in \emph{Proc. 12th Int. Conf. Doc. Anal.
  Recognit.}, 2013, pp. 1484--1493.

\bibitem{chen_textnaturalscenes_2004}
{Xiangrong Chen} and A.~Yuille, ``{Detecting and reading text in natural
  scenes},'' in \emph{Proc. IEEE Comput. Soc. Conf. Comput. Vis. Pattern
  Recognit.}, vol.~2, 2004, pp. 366--373.

\bibitem{Neumann_textlocandrec_2011}
L.~Neumann and J.~Matas, ``{A method for text localization and recognition in
  real-world images},'' in \emph{Proc. Asian Conf. Comput. Vis.}, vol. 6494,
  2011, pp. 770--783.

\bibitem{jamil_edge-based_2011}
A.~Jamil, I.~Siddiqi, F.~Arif, and A.~Raza, ``{Edge-Based Features for
  Localization of Artificial Urdu Text in Video Images},'' in \emph{Proc. Int.
  Conf. Doc. Anal. Recognit.}, 2011, pp. 1120--1124.

\bibitem{mosleh_automatic_2013}
A.~Mosleh, N.~Bouguila, and A.~B. Hamza, ``{Automatic inpainting scheme for
  video text detection and removal},'' \emph{IEEE Trans. Image Process.},
  vol.~22, no.~11, pp. 4460--4472, 2013.

\bibitem{huang_textloc_2013}
W.~Huang, Z.~Lin, J.~Yang, and J.~Wang, ``{Text Localization in Natural Images
  Using Stroke Feature Transform and Text Covariance Descriptors},'' in
  \emph{Proc. IEEE Int. Conf. Comput. Vis.}, 2013, pp. 1241--1248.

\bibitem{Liaotextbox2017}
M.~Liao, B.~Shi, X.~Bai, X.~Wang, and W.~Liu, ``{TextBoxes: A fast text
  detector with a single deep neural network},'' in \emph{Proc. AAAI Conf.
  Artif. Intell.}, 2017, pp. 4161--4167.

\bibitem{liu_ssd_2016}
W.~Liu, D.~Anguelov, D.~Erhan, C.~Szegedy, S.~Reed, C.~Y. Fu, and A.~C. Berg,
  ``{SSD: Single shot multibox detector},'' in \emph{Proc. Eur. Conf. Comput.
  Vis.}, vol. 9905, 2016, pp. 21--37.

\bibitem{tian_CTPN_2016}
Z.~Tian, W.~Huang, T.~He, P.~He, and Y.~Qiao, ``{Detecting text in natural
  image with connectionist text proposal network},'' in \emph{Proc. Eur. Conf.
  Comput. Vis.}, vol. 9912, 2016, pp. 56--72.

\bibitem{RenfasterRCNN2015}
S.~Ren, K.~He, R.~Girshick, and J.~Sun, ``{Faster R-CNN: Towards real-time
  object detection with region proposal networks},'' in \emph{Proc. Adv. Neural
  Inf. Process. Syst.}, 2015, pp. 91--99.

\bibitem{ma_RRPN_2018}
J.~Ma, W.~Shao, H.~Ye, L.~Wang, H.~Wang, Y.~Zheng, and X.~Xue,
  ``{Arbitrary-Oriented Scene Text Detection via Rotation Proposals},''
  \emph{IEEE Trans. Multimed.}, vol.~20, no.~11, pp. 3111--3122, 2018.

\bibitem{zhou_east_2017}
X.~Zhou, C.~Yao, H.~Wen, Y.~Wang, S.~Zhou, W.~He, and J.~Liang, ``{EAST: An
  Efficient and Accurate Scene Text Detector},'' in \emph{Proc. IEEE Conf.
  Comput. Vis. Pattern Recognit.}, 2017, pp. 2642--2651.

\bibitem{zhang_look_2019}
C.~Zhang, B.~Liang, Z.~Huang, M.~En, J.~Han, E.~Ding, and X.~Ding, ``{Look More
  Than Once: An Accurate Detector for Text of Arbitrary Shapes},'' in
  \emph{Proc. IEEE/CVF Conf. Comput. Vis. Pattern Recognit.}, 2019, pp.
  10\,544--10\,553.

\bibitem{ZhangMutidetectFCN2016}
Z.~Zhang, C.~Zhang, W.~Shen, C.~Yao, W.~Liu, and X.~Bai, ``{Multi-oriented Text
  Detection with Fully Convolutional Networks},'' in \emph{Proc. IEEE Conf.
  Comput. Vis. Pattern Recognit.}, 2016, pp. 4159--4167.

\bibitem{Lyumasktextspotter2018}
P.~Lyu, M.~Liao, C.~Yao, W.~Wu, and X.~Bai, ``{Mask textspotter: An end-to-end
  trainable neural network for spotting text with arbitrary shapes},'' in
  \emph{Proc. Eur. Conf. Comput. Vis.}, vol. 11218, 2018, pp. 71--88.

\bibitem{he_maskRCNN_2017}
K.~He, G.~Gkioxari, P.~Dollar, and R.~Girshick, ``{Mask R-CNN},'' in
  \emph{Proc. IEEE Int. Conf. Comput. Vis.}, 2017, pp. 2980--2988.

\bibitem{long_textsnake_2018}
S.~Long, J.~Ruan, W.~Zhang, X.~He, W.~Wu, and C.~Yao, ``{TextSnake: A Flexible
  Representation for Detecting Text of Arbitrary Shapes},'' in \emph{Proc. Eur.
  Conf. Comput. Vis.}, vol. 11206, 2018, pp. 19--35.

\bibitem{LiPSENet2020}
Y.~Li, Z.~Wu, S.~Zhao, X.~Wu, Y.~Kuang, Y.~Yan, S.~Ge, K.~Wang, W.~Fan,
  X.~Chen, and Y.~Wang, ``{PSENet: Psoriasis Severity Evaluation Network},'' in
  \emph{Proc. AAAI Conf. Artif. Intell.}, vol.~34, no.~01, 2020, pp. 800--807.

\bibitem{LiaoDB2020}
M.~Liao, Z.~Wan, C.~Yao, K.~Chen, and X.~Bai, ``{Real-Time Scene Text Detection
  with Differentiable Binarization},'' in \emph{Proc. AAAI Conf. Artif.
  Intell.}, vol.~34, no.~07, 2020, pp. 11\,474--11\,481.

\bibitem{BaekCRAFT2019}
Y.~Baek, B.~Lee, D.~Han, S.~Yun, and H.~Lee, ``{Character Region Awareness for
  Text Detection},'' in \emph{Proc. IEEE/CVF Conf. Comput. Vis. Pattern
  Recognit.}, 2019, pp. 9357--9366.

\bibitem{efros_image_2001}
A.~A. Efros and W.~T. Freeman, ``{Image quilting for texture synthesis and
  transfer},'' in \emph{Proc. 28th Annu. Conf. Comput. Graph. Interact. Tech. -
  SIGGRAPH '01}, 2001, pp. 341--346.

\bibitem{barnes_patchmatch_2009}
C.~Barnes, E.~Shechtman, A.~Finkelstein, and D.~B. Goldman, ``{PatchMatch},''
  \emph{ACM Trans. Graph.}, vol.~28, no.~3, pp. 1--11, 2009.

\bibitem{darabi_inpaintpatch_2012}
S.~Darabi, E.~Shechtman, C.~Barnes, {Dan B Goldman}, and P.~Sen, ``{Image
  melding: Combining inconsistent images using patch-based synthesis},''
  \emph{ACM Trans. Graph.}, vol.~31, no.~4, pp. 1--10, 2012.

\bibitem{Bertalmio2000}
M.~Bertalmio, G.~Sapiro, V.~Caselles, and C.~Ballester, ``{Image inpainting},''
  in \emph{Proc. ACM SIGGRAPH Conf. Comput. Graph.}, 2000.

\bibitem{Oliveira2001}
M.~M. Oliveira, B.~Bowen, R.~McKenna, and Y.-S. Chang, ``{Fast Digital Image
  Inpainting},'' \emph{Int. Conf. Vis. Imaging Image Process.}, 2001.

\bibitem{Pathakcontextencoder2016}
D.~Pathak, P.~Krahenbuhl, J.~Donahue, T.~Darrell, and A.~A. Efros, ``{Context
  Encoders: Feature Learning by Inpainting},'' in \emph{Proc. IEEE Conf.
  Comput. Vis. Pattern Recognit.}, 2016, pp. 2536--2544.

\bibitem{iizuka_globally_2017}
S.~Iizuka, E.~Simo-Serra, and H.~Ishikawa, ``{Globally and locally consistent
  image completion},'' \emph{ACM Trans. Graph.}, vol.~36, no.~4, pp. 1--14,
  2017.

\bibitem{liu_partialConv_2018}
G.~Liu, F.~A. Reda, K.~J. Shih, T.~C. Wang, A.~Tao, and B.~Catanzaro, ``{Image
  Inpainting for Irregular Holes Using Partial Convolutions},'' in \emph{Proc.
  Eur. Conf. Comput. Vis.}, vol. 11215, 2018, pp. 89--105.

\bibitem{xie_LBAM_2019}
C.~Xie, S.~Liu, C.~Li, M.-M. Cheng, W.~Zuo, X.~Liu, S.~Wen, and E.~Ding,
  ``{Image Inpainting With Learnable Bidirectional Attention Maps},'' in
  \emph{Proc. IEEE/CVF Int. Conf. Comput. Vis.}, 2019, pp. 8857--8866.

\bibitem{li_recurrent_2020}
J.~Li, N.~Wang, L.~Zhang, B.~Du, and D.~Tao, ``{Recurrent Feature Reasoning for
  Image Inpainting},'' in \emph{Proc. IEEE/CVF Conf. Comput. Vis. Pattern
  Recognit.}, 2020, pp. 7757--7765.

\bibitem{Yu_ContextualAttention_2018}
J.~Yu, Z.~Lin, J.~Yang, X.~Shen, X.~Lu, and T.~S. Huang, ``{Generative Image
  Inpainting with Contextual Attention},'' in \emph{Proc. IEEE/CVF Conf.
  Comput. Vis. Pattern Recognit.}, 2018, pp. 5505--5514.

\bibitem{yu_gatedConv_2019}
J.~Yu, Z.~Lin, J.~Yang, X.~Shen, X.~Lu, and T.~Huang, ``{Free-Form Image
  Inpainting With Gated Convolution},'' in \emph{Proc. IEEE/CVF Int. Conf.
  Comput. Vis.}, 2019, pp. 4470--4479.

\bibitem{yi_HiFill_2020}
Z.~Yi, Q.~Tang, S.~Azizi, D.~Jang, and Z.~Xu, ``{Contextual Residual
  Aggregation for Ultra High-Resolution Image Inpainting},'' in \emph{Proc.
  IEEE/CVF Conf. Comput. Vis. Pattern Recognit.}, 2020, pp. 7505--7514.

\bibitem{Leetextremovalvideo2003}
C.~W. Lee, K.~Jung, and H.~J. Kim, ``{Automatic text detection and removal in
  video sequences},'' \emph{Pattern Recognit. Lett.}, vol.~24, no.~15, pp.
  2607--2623, 2003.

\bibitem{Pnevmatikakis2008}
E.~A. Pnevmatikakis and P.~Maragos, ``{An inpainting system for automatic image
  structure - texture restoration with text removal},'' in \emph{Proc. 15th
  IEEE Int. Conf. Image Process.}, 2008, pp. 2616--2619.

\bibitem{mosleh_texterasing_2012}
A.~Mosleh, N.~Bouguila, and A.~B. Hamza, ``{Image Text Detection Using a
  Bandlet-Based Edge Detector and Stroke Width Transform},'' in \emph{Proc. Br.
  Mach. Vis. Conf.}, 2012, pp. 63.1--63.12.

\bibitem{Khodadadi2012}
M.~Khodadadi and A.~Behrad, ``{Text localization, extraction and inpainting in
  color images},'' in \emph{Proc. 20th Iran. Conf. Electr. Eng.}, 2012, pp.
  1035--1040.

\bibitem{wagh_textremoval_2015}
P.~D. Wagh and D.~R. Patil, ``{Text detection and removal from image using
  inpainting with smoothing},'' in \emph{Proc. Int. Conf. Pervasive Comput.},
  2015.

\bibitem{Isolapixel2pixel2017}
P.~Isola, J.-Y. Zhu, T.~Zhou, and A.~A. Efros, ``{Image-to-Image Translation
  with Conditional Adversarial Networks},'' in \emph{Proc. IEEE Conf. Comput.
  Vis. Pattern Recognit.}, 2017, pp. 5967--5976.

\bibitem{QinBMVC2018}
S.~Qin, J.~Wei, and R.~Manduchi, ``{Automatic semantic content removal by
  learning to neglect},'' in \emph{Proc. Br. Mach. Vis. Conf.}, 2018.

\bibitem{ZhengPIC2019}
C.~Zheng, T.-J. Cham, and J.~Cai, ``{Pluralistic Image Completion},'' in
  \emph{Proc. IEEE/CVF Conf. Comput. Vis. Pattern Recognit.}, 2019, pp.
  1438--1447.

\bibitem{HeResNet2016}
K.~He, X.~Zhang, S.~Ren, and J.~Sun, ``{Deep Residual Learning for Image
  Recognition},'' in \emph{Proc. IEEE Conf. Comput. Vis. Pattern Recognit.},
  2016, pp. 770--778.

\bibitem{Milletariunet2016}
F.~Milletari, N.~Navab, and S.-A. Ahmadi, ``{V-Net: Fully Convolutional Neural
  Networks for Volumetric Medical Image Segmentation},'' in \emph{Proc. Int.
  Conf. 3D Vis.}, 2016, pp. 565--571.

\bibitem{cao_GCNet_2019}
Y.~Cao, J.~Xu, S.~Lin, F.~Wei, and H.~Hu, ``{GCNet: Non-Local Networks Meet
  Squeeze-Excitation Networks and Beyond},'' in \emph{Proc. IEEE/CVF Int. Conf.
  Comput. Vis. Work.}, 2019, pp. 1971--1980.

\bibitem{johnson_perceptual_2016}
J.~Johnson, A.~Alahi, and L.~Fei-Fei, ``{Perceptual losses for real-time style
  transfer and super-resolution},'' in \emph{Proc. Eur. Conf. Comput. Vis.},
  vol. 9906, 2016, pp. 694--711.

\bibitem{Gatysstyle2016}
L.~A. Gatys, A.~S. Ecker, and M.~Bethge, ``{Image Style Transfer Using
  Convolutional Neural Networks},'' in \emph{Proc. IEEE Conf. Comput. Vis.
  Pattern Recognit.}, 2016, pp. 2414--2423.

\bibitem{SimonyanVGG2015}
K.~Simonyan and A.~Zisserman, ``{Very deep convolutional networks for
  large-scale image recognition},'' in \emph{Proc. 3rd Int. Conf. Learn.
  Represent.}, 2015.

\bibitem{KingmaAdam2015}
D.~P. Kingma and J.~L. Ba, ``{Adam: A method for stochastic optimization},'' in
  \emph{Proc. 3rd Int. Conf. Learn. Represent.}, 2015.

\bibitem{NayefICDAR2017MLT}
N.~Nayef, F.~Yin, I.~Bizid, H.~Choi, Y.~Feng, D.~Karatzas, Z.~Luo, U.~Pal,
  C.~Rigaud, J.~Chazalon, W.~Khlif, M.~M. Luqman, J.-C. Burie, C.-l. Liu, and
  J.-M. Ogier, ``{ICDAR2017 Robust Reading Challenge on Multi-Lingual Scene
  Text Detection and Script Identification - RRC-MLT},'' in \emph{Proc. 14th
  IAPR Int. Conf. Doc. Anal. Recognit.}, vol.~1, 2017, pp. 1454--1459.

\bibitem{KaratzasICDAR2015}
D.~Karatzas, L.~Gomez-Bigorda, A.~Nicolaou, S.~Ghosh, A.~Bagdanov, M.~Iwamura,
  J.~Matas, L.~Neumann, V.~R. Chandrasekhar, S.~Lu, F.~Shafait, S.~Uchida, and
  E.~Valveny, ``{ICDAR 2015 competition on Robust Reading},'' in \emph{Proc.
  13th Int. Conf. Doc. Anal. Recognit.}, 2015, pp. 1156--1160.

\bibitem{veit_cocotext_2016}
A.~Veit, T.~Matera, L.~Neumann, J.~Matas, and S.~Belongie, ``{COCO-Text:
  Dataset and Benchmark for Text Detection and Recognition in Natural
  Images},'' \emph{arXiv}, 2016.

\bibitem{Wangwordspot2010}
K.~Wang and S.~Belongie, ``{Word spotting in the wild},'' in \emph{Proc. Eur.
  Conf. Comput. Vis.}, vol. 6311, 2010, pp. 591--604.

\bibitem{NayefICDAR2019MLT}
N.~Nayef, C.-l. Liu, J.-M. Ogier, Y.~Patel, M.~Busta, P.~N. Chowdhury,
  D.~Karatzas, W.~Khlif, J.~Matas, U.~Pal, and J.-C. Burie, ``{ICDAR2019 Robust
  Reading Challenge on Multi-lingual Scene Text Detection and Recognition —
  RRC-MLT-2019},'' in \emph{Proc. Int. Conf. Doc. Anal. Recognit.}, 2019, pp.
  1582--1587.

\bibitem{ChngICDAR2019ArT}
C.~K. Chng, E.~Ding, J.~Liu, D.~Karatzas, C.~S. Chan, L.~Jin, Y.~Liu, Y.~Sun,
  C.~C. Ng, C.~Luo, Z.~Ni, C.~Fang, S.~Zhang, and J.~Han, ``{ICDAR2019 Robust
  Reading Challenge on Arbitrary-Shaped Text - RRC-ArT},'' in \emph{Proc. Int.
  Conf. Doc. Anal. Recognit.}, 2019, pp. 1571--1576.

\bibitem{WangSSIM2004}
Z.~Wang, A.~C. Bovik, H.~R. Sheikh, and E.~P. Simoncelli, ``{Image quality
  assessment: From error visibility to structural similarity},'' \emph{IEEE
  Trans. Image Process.}, vol.~13, no.~4, pp. 600--612, 2004.

\bibitem{zhou_places_2018}
B.~Zhou, A.~Lapedriza, A.~Khosla, A.~Oliva, and A.~Torralba, ``{Places: A 10
  Million Image Database for Scene Recognition},'' \emph{IEEE Trans. Pattern
  Anal. Mach. Intell.}, vol.~40, no.~6, pp. 1452--1464, 2018.

\end{thebibliography}


% \end{thebibliography}


\begin{IEEEbiography}[{\includegraphics[width=1in,height=1.25in,clip,keepaspectratio]{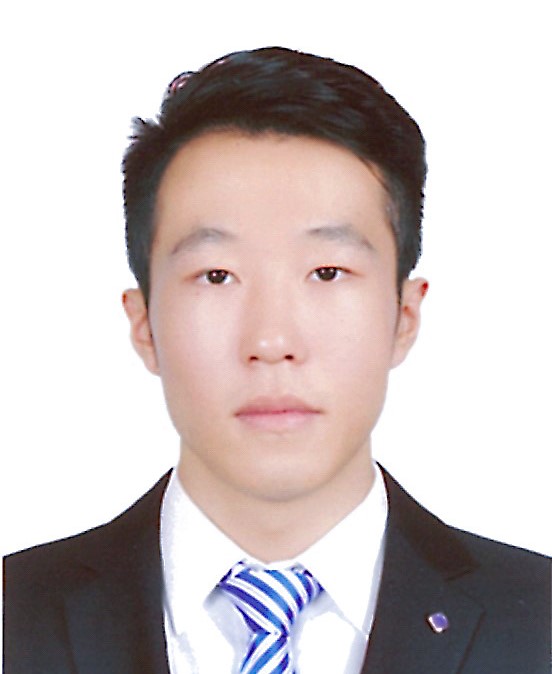}}]{Zhengmi Tang}
received his B.E. degree from Xidian University, Shaanxi, China, in 2017 and his M.E. degree in cybernetics engineering from Hiroshima University, Japan, in 2020. He is currently pursuing a Ph.D. degree in communication engineering at the IIC-Lab at Tohoku University, Japan. His current research interests include computer vision, scene-text detection, and data synthesis.
\end{IEEEbiography}

\begin{IEEEbiography}[{\includegraphics[width=1in,height=1.25in,clip,keepaspectratio]{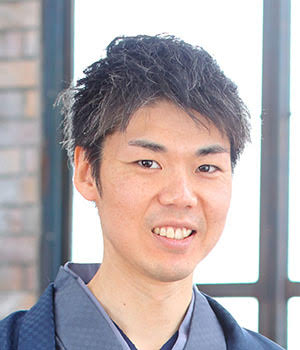}}]{Tomo Miyazaki}
 (Member, IEEE) received his B.E. and Ph.D. degrees from Yamagata University (2006) and Tohoku University (2011), respectively. From 2011 to 2012, he worked on the geographic information system at Hitachi, Ltd. From 2013 to 2014, he worked at Tohoku University as a postdoctoral researcher. Since 2015, he has been an Assistant Professor at the university. His research interests include pattern recognition and image processing.
\end{IEEEbiography}

\begin{IEEEbiography}[{\includegraphics[width=1in,height=1.25in,clip,keepaspectratio]{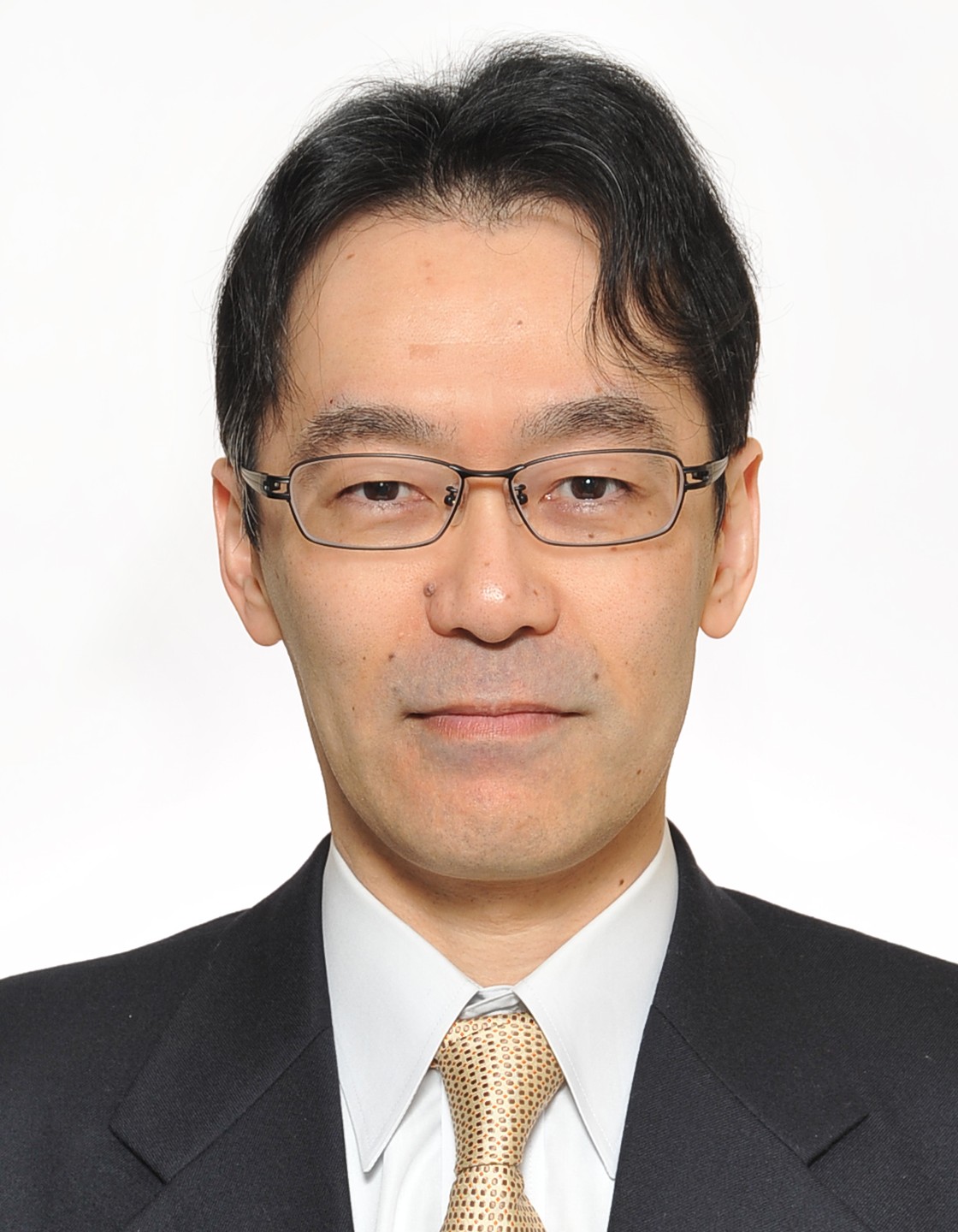}}]{Yoshihiro Sugaya}
(Member, IEEE) received his B.E., M.E., and Ph.D. degrees from Tohoku University, Sendai, Japan in 1995, 1997, and 2002, respectively. He is currently an Associate Professor at the Graduate School of Engineering, Tohoku University. His research interests include computer vision, pattern recognition, image processing, parallel processing, and distributed computing. Dr. Sugaya is a member of the Institute of Electronics, Information and Communication Engineers (IEICE) and the Information Processing Society of Japan.
\end{IEEEbiography}

%\newpage

\begin{IEEEbiography}[{\includegraphics[width=1in,height=1.25in,clip,keepaspectratio]{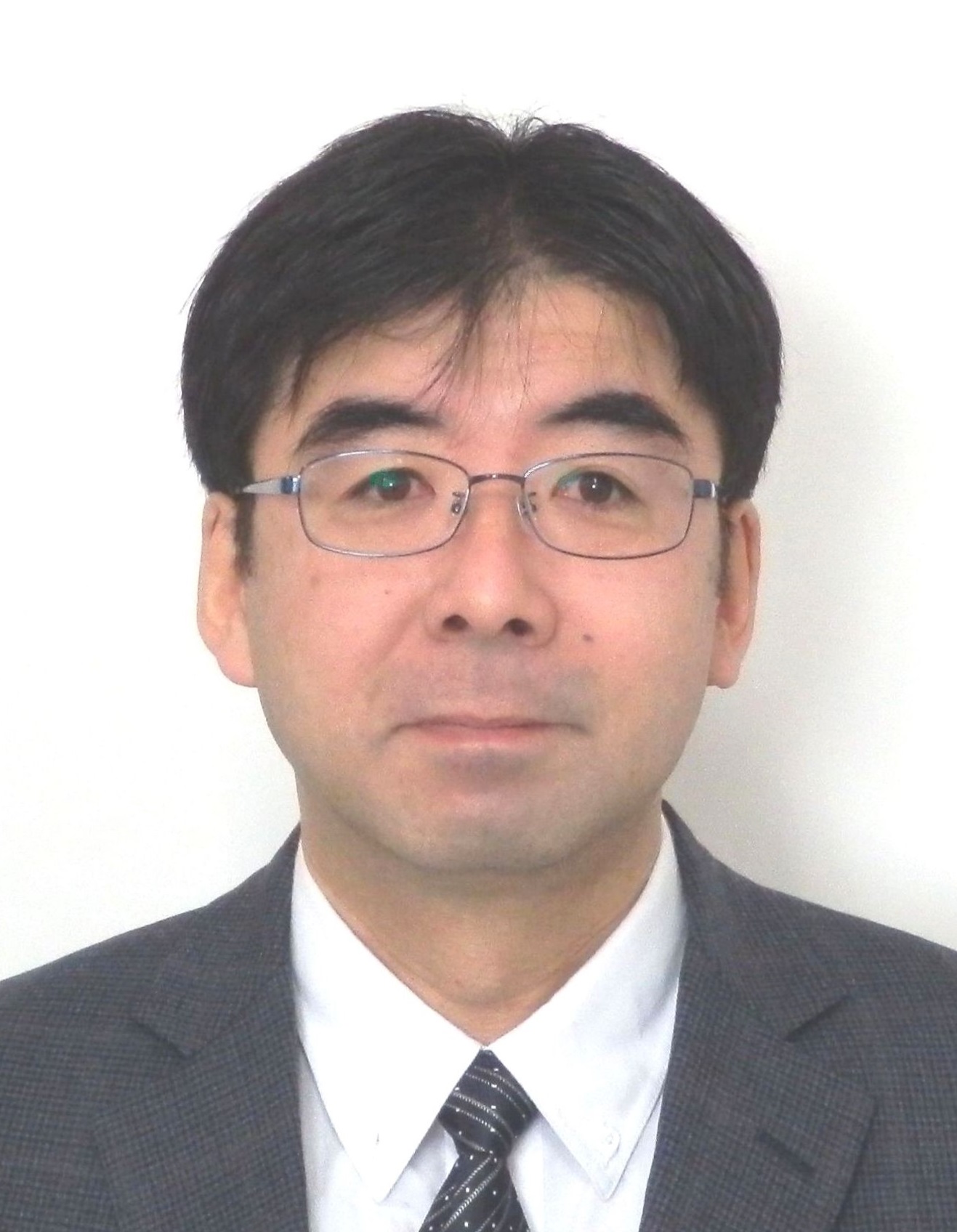}}]{Shinichiro Omachi}
(M’96-SM’11) received his B.E., M.E., and Ph.D. degrees in Information Engineering from Tohoku University, Japan, in 1988, 1990, and 1993, respectively. He worked as an Assistant Professor at the Education Center for Information Processing at Tohoku University from 1993 to 1996. Since 1996, he has been affiliated with the Graduate School of Engineering at Tohoku University, where he is currently a Professor. From 2000 to 2001, he was a visiting Associate Professor at Brown University. His research interests include pattern recognition, computer vision, image processing, image coding, and parallel processing. He served as the Editor-in-Chief of IEICE Transactions on Information and Systems from 2013 to 2015. Dr. Omachi is a member of the Institute of Electronics, Information and Communication Engineers, the Information Processing Society of Japan, among others. He received the IAPR/ICDAR Best Paper Award in 2007, the Best Paper Method Award of the 33rd Annual Conference of the GfKl in 2010, the ICFHR Best Paper Award in 2010, and the IEICE Best Paper Award in 2012. He is currently the Vice Chair of the IEEE Sendai Section.
\end{IEEEbiography}


%\vfill


% that', s all folks
\end{document}